\documentclass{article} %
\usepackage{iclr2027_conference,times}
\usepackage[T1]{fontenc}
\usepackage{pifont}
\usepackage{ulem}
\usepackage{algorithm} 
\usepackage[noend]{algpseudocode}

\usepackage{amsmath,amsfonts,bm}

\def\eqref#1{equation~\ref{#1}}

\def\1{\bm{1}}

\DeclareMathAlphabet{\mathsfit}{\encodingdefault}{\sfdefault}{m}{sl}
\SetMathAlphabet{\mathsfit}{bold}{\encodingdefault}{\sfdefault}{bx}{n}

\newcommand{\Unif}{\operatorname{Unif}}

\newcommand{\stopk}{\mathsf{stop}}
\newcommand{\rolek}{\mathsf{role}}

\newcommand{\edgek}{\mathsf{edge}}
\newcommand{\ind}{\mathbb{I}}

\DeclareMathOperator*{\argmax}{arg\,max}

\usepackage{listings}
\usepackage[most]{tcolorbox}

\definecolor{roleframe}{HTML}{B0AEA5}   %
\definecolor{roletitle}{HTML}{E8E6DC}   %
\definecolor{roleink}{HTML}{141413}     %

\lstdefinestyle{roleprofile}{
  basicstyle=\ttfamily\footnotesize\color{roleink},
  breaklines=true, breakindent=3em,
  columns=fullflexible, keepspaces=true,
  upquote=true, showstringspaces=false,
  morekeywords={Name,Description,Prompt,Tool,Reads},
  keywordstyle=\bfseries,
}

\newtcblisting{roleprofile}[1]{
  listing only, listing options={style=roleprofile},
  breakable, enhanced,
  title={\textbf{Role profile}\hfill #1},
  fonttitle=\small, coltitle=roleink, colbacktitle=roletitle,
  colframe=roleframe, colback=white,
  boxrule=0.4pt, arc=0pt,
  left=8pt, right=8pt, top=4pt, bottom=4pt,
  before skip=10pt, after skip=10pt,
}

\usepackage{hyperref}
\usepackage{url}
\usepackage{graphicx}
\usepackage{booktabs}
\usepackage{multirow}
\usepackage[table]{xcolor}
\usepackage{wrapfig}
\usepackage{float}
\usepackage{placeins}
\usepackage{longtable}
\usepackage{array}
\graphicspath{{./}{../}}
\makeatletter\def\input@path{{./}{../}}\makeatother
\newcommand{\cmk}{\textcolor{green!60!black}{\ding{51}}}
\newcommand{\xmk}{\textcolor{red!70!black}{\ding{55}}}
\providecommand{\bytes}[1]{\lvert #1\rvert_{\mathrm{B}}}
\providecommand{\Enc}{\operatorname{Enc}}
\providecommand{\MLP}{\operatorname{MLP}}

\providecommand{\LN}{\operatorname{LN}}
\providecommand{\Ind}{\mathbf{1}}

\providecommand{\clip}{\operatorname{clip}}
\title{MeterRouter: Budget-Aware Dynamic Routing for Multi-Agent Systems}

\author{
\textbf{Yue Xu}$^{1,2,\dagger}$ \quad
\textbf{Leikun Yin}$^{2,\dagger}$ \quad
\textbf{Ruiming Min}$^{3,\dagger}$ \quad
\textbf{Yufeng Lai}$^{4}$ \quad
\textbf{Justin Andrew Johnson}$^{1}$ \\
\textbf{Sitao Luan}$^{5,6,*}$ \quad
\textbf{Xiaotong Shen}$^{3,*}$\\[0.5em]
$^{1}$Department of Applied Economics, University of Minnesota\\
$^{2}$Department of Computer Science and Engineering, University of Minnesota\\
$^{3}$School of Statistics, University of Minnesota\\
$^{4}$Databricks\\
$^{5}$Mila -- Quebec Artificial Intelligence Institute \\
$^{6}$University of Montreal\\[0.5em]
$^{\dagger}$Equal contribution \quad $^{*}$Corresponding authors
}

\iclrfinalcopy %
\begin{document}

\maketitle
\lhead{} %
\renewcommand{\headrulewidth}{0pt}
{\renewcommand{\thefootnote}{\fnsymbol{footnote}}%
\footnotetext[2]{Emails: Yue Xu \texttt{<xu001302@umn.edu>}, Leikun Yin \texttt{<yin00406@umn.edu>}, Ruiming Min \texttt{<min00089@umn.edu>}.}%
}

\begin{abstract}

Multi-agent systems (MAS) built on large language models (LLMs) outperform single-agent systems, but inter-agent communication incurs substantial token overhead. While assigning a large LLM to every agent is common, it results in inefficient resource utilization, leaving the system far from the Pareto frontier. The cost of answering a query is set jointly by how many agents are recruited, the role and LLM assigned to each, and communication topology among them. Most existing routers treat cost only implicitly rather than as an explicit optimization constraint. A fixed accuracy–cost preference also cannot express how much a caller is willing to pay for a particular query. We introduce \texttt{MeterRouter}, a budget-aware dynamic router that incrementally synthesizes a collaboration graph by selecting each agent's role, LLM, incoming connections, and termination conditioned on the query and a per-query monetary budget. We initialize the router by imitating successful execution graphs, then jointly refine its decisions with a variant of Group Relative Policy Optimization (GRPO) whose reward favors correct answers and penalizes spending and budget overruns. Across five benchmarks and a three-tier LLM pool, \texttt{MeterRouter} is \textbf{(1) accurate}, achieving performance comparable to the strongest MAS; \textbf{(2) economical}, reaching that accuracy at 82\% lower cost than the strongest MAS router and extending the accuracy--cost frontier beyond the ceiling of single-agent systems; and \textbf{(3) controllable}, adapting its spending to the specified budget while rarely exceeding it.

\end{abstract}

\section{Introduction}

Large Language Model (LLM)-based agents have demonstrated remarkable capabilities in tackling complex tasks that extend beyond enclosed reasoning and textual boundaries \citep{toolformer}, with external capabilities like context-aware memory \citep{treacle}, tool use \citep{toolllm}. However, single-agent architectures often struggle when executing multi-step reasoning \citep{cot,planningabilities}, adapting to dynamic environments \citep{react}, or maintaining coherence over extended horizons \citep{generativeagents, contextfolding}. To address these challenges, Multi-Agent Systems (MAS) have emerged as a dominant paradigm, consistently outperforming single-agent by well-crafted inter-agent communication topologies, harnessing collective intelligence \citep{agentverse, moa}, specialized role allocation \citep{metagpt,chatdev}, and dynamic error recovery \citep{reflexion,llmdebate,critic}. 

With the rapid expansion of a highly accessible and heterogeneous LLM landscape, selecting appropriate LLMs based on their capability and cost %
has increasingly captured the attention of the research community~\citep{masrouter, bamas, agentbalance}.
Although LLM capabilities generally scale with model size, smaller models reduce computational cost \citep{graphrouter}.
Moreover, uniform dependency on a single model introduces critical operational vulnerabilities, leading to service outages or demand spikes \citep{frugalgpt}.
 LLM-routing is proposed to assign the suitable LLM for each query~\citep{c2mabv,hybridllm, routellm,graphrouter,agentrouter}.
Leveraging model diversity by strategically assigning heterogeneous LLMs offers a promising approach for MAS construction.

MAS cooperative performance gains rely on inter-agent information transmission, exchange and assimilation \citep{agentprune}. Such extensive prompt-level communication-spanning inter-agent and self-evaluation introduces substantial token overhead \citep{tokeneconomies}, leading to a fundamental trade-off between collaborative synergy and token efficiency. To address that, the MAS topology design has shifted from static, manually designed graphs to more flexible and sparse task-adaptive topology synthesis paradigms~\citep{agentdropout,gdesigner,argdesigner}. However, these cost-efficiency solutions operate exclusively at the structural orchestration level, uniformly assigning homogeneous LLMs without leveraging heterogeneous LLMs. 
This motivates a critical question: \textit{How to architect a budget-aware, topologically adaptive MAS using heterogeneous backbones that optimizes task performance with cost-efficiency}?

\begin{figure*}[t]
    \centering
    \includegraphics[width=1.00\textwidth]{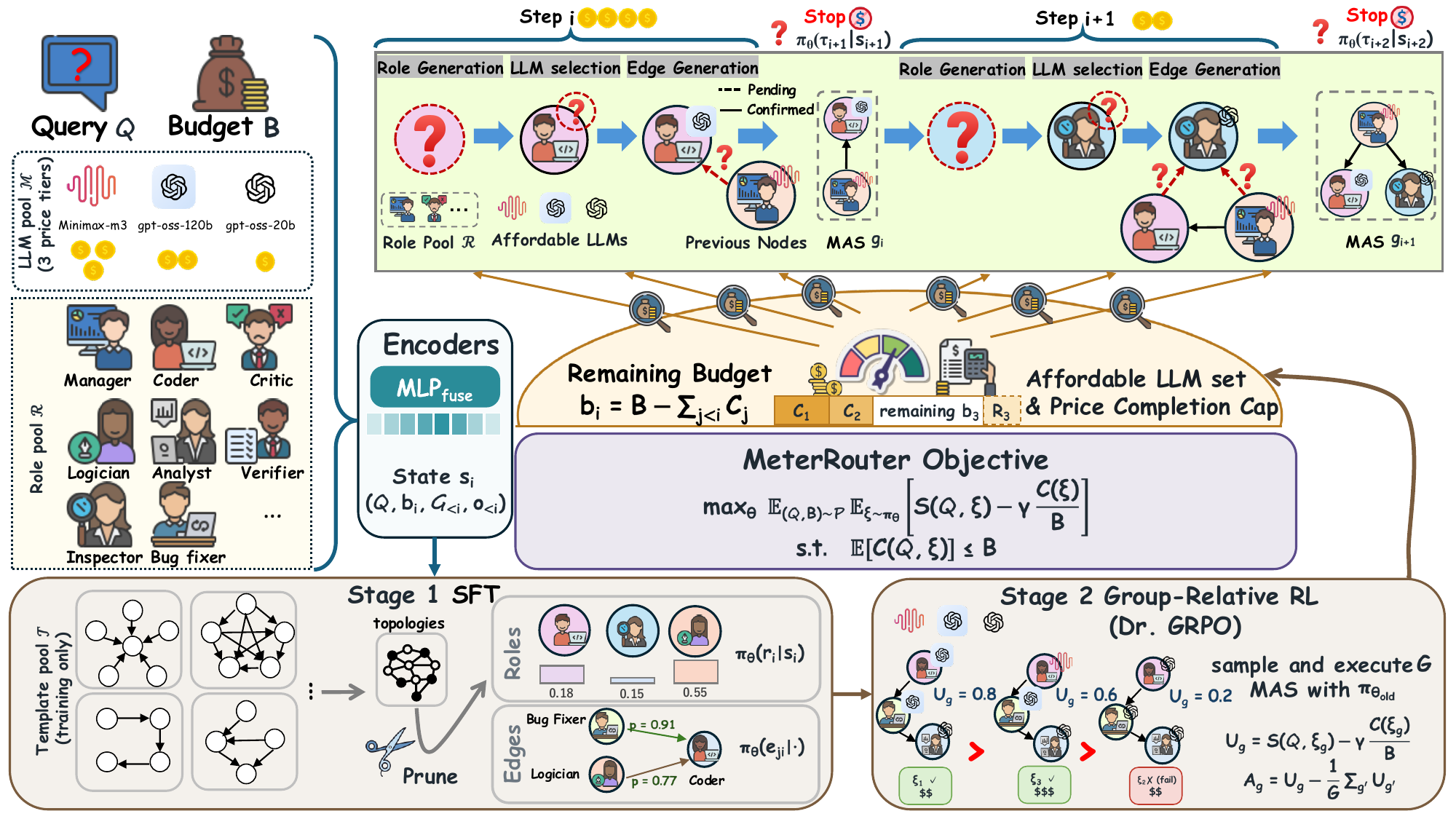}
    \caption{\textbf{Overview of \texttt{MeterRouter}.} The router builds the MAS incrementally, choosing each agent's role, backbone, and incoming edges, or stopping, under the remaining budget tracked by the meter (top, middle). It is trained by budget-conditioned imitation (Stage 1) and group-relative RL (Stage 2) (bottom).}
    \label{fig:overview}
\end{figure*}

For complex tasks where the division of labor is hard to define, fixed communication topologies including chains
 cannot meaningfully contribute to collaborative intelligence~\citep{agentprune}. In contrast, sparse, task-adaptive graphs reduce redundant roles and edges~\citep{gdesigner,argdesigner}. However, designing these topologies independently of model selection can still lead to suboptimal MAS construction. To avoid separating those two selections, we propose \texttt{MeterRouter} (Figure~\ref{fig:overview}), a framework that %
jointly optimizes topology generation and LLM-to-role assignments. By dynamically tailoring agent topologies \citep{argdesigner} and LLM assignments by remaining budget, \texttt{MeterRouter} optimizes the accuracy-cost tradeoff, achieving competitive task performance while substantially reducing monetary consumption. Our main contributions are as follows:
\ding{202} \textbf{Autonomous MAS Orchestration with Heterogeneous LLMs:} We design MAS routing, which dynamic topology generation with assigning each agent’s role, backbone, and incoming edges, systematically eliminates the operational redundancies and suboptimal LLM assignments inherent in static, pre-defined templates.
\ding{203} \textbf{Budget-Aware and Dynamic Adjustment:} We introduce a flexible and practical multi-agent framework customized to user-defined budget per query. Instead of cut-off output generation, \texttt{MeterRouter} employs a dynamic remaining-budget tracker to adaptively reallocate resources to achieve cost-efficient
routing.
\ding{204} \textbf{Pareto Optimization:} We introduce \texttt{MeterRouter}, a joint routing mechanism that balances task performance and token cost. Driven by a dual-source cost reduction strategy, it minimizes operational overhead by combining structural communication savings with heterogeneous capacity assignment savings.

\section{Related works}
\noindent\textbf{Collaboration Topology.}
MAS have transitioned from fixed interaction heuristics to topology optimization. The progression occurs along two critical axes: \textbf{task adaptivity} and \textbf{structural flexibility}. Early frameworks relied on static canonical graphs, such as CoT \citep{cot}, Reflexion \citep{reflexion}, LLM-debate \citep{llmdebate} and MACNET \citep{macnet}. These networks fundamentally dictate reasoning performance while remaining task-agnostic and computationally inefficient. Dynamic task-adaptive frameworks advance beyond rigid heuristics by leveraging execution feedback to steer communication traces~\citep{metagpt, puppets,autogen}. 
To eliminate manual topology engineering, template-based optimization frameworks apply learnable edge reweighting and pruning mechanisms to distill redundant communication topologies into sparse structures \citep{gptswarm,agentdropout,agentprune}. Pushing beyond template modification, graph generative paradigms construct autonomous topologies: G-Designer \citep{gdesigner} utilizes a graph autoencoder to learn task-adaptive collaboration structures, while ARG-Designer \citep{argdesigner} leverages an autoregressive model to incrementally build %
graphs. 
However, existing MAS frameworks frequently over-provision computational resources by uniformly assigning top-tier LLMs to all agents, leading to excessive token consumption \citep{masrouter}. 

\noindent\textbf{LLM Routing.}
LLM routing optimizes the Pareto frontier between cost and response quality by assigning the suitable LLMs for queries. Early paradigms rely on sequential cascading, such as FrugalGPT \citep{frugalgpt}, which queries models iteratively until satisfying a confidence threshold, or binary routing based on predicted query difficulty \citep{hybridllm} and human preference \citep{routellm}. Recognizing that ensembling responses across LLMs outperforms a single LLM, recent works transition from binary model selection to multi-choice routing and output fusion~\citep{c2mabv, routerdc, graphrouter, agentrouter}. %
Extending routing to MAS, MasRouter \citep{masrouter} integrates topology selection, role assignment and LLM routing, but it is constrained to selecting from a graph catalog. ProgRouter \citep{progrouter} performs step-wise LLM routing across workflow pipelines. %

\section{Problem Formulation}
\subsection{Notation Establishment}
\noindent\textbf{Search Space.} $\mathcal{R}$ denotes the pool of available $N_r$ roles (e.g., programmer, analyst), $\mathcal{M}$ denotes $N_m$ available LLM backbone pool, and each LLM backbone carries a  price pair
$(c^{\mathrm{in}}_m,c^{\mathrm{out}}_m)$ in dollars per input and output token.  More detailed notation form is in Appendix~\ref{app:notation}.

\noindent\textbf{Agents and edges.}
We represent a multi-agent system as a directed acyclic graph (DAG)
$\mathcal{G} = (\mathcal{V}, \mathcal{E})$. The node set $\mathcal{V} = \{v_1, v_2, \cdots, v_N\}$ represents collection of $N$ interacting agents and agent $v_i$ is an instance with an LLM backbone $m_i$ endowed with a specific role function and expertise $r_i$ denoted as
\(
v_i=\{r_i,m_i\}, r_i\in\mathcal{R}, m_i\in \mathcal{M}.
\)
Edges $\mathcal{E} \subseteq \mathcal{V} \times \mathcal{V}$ represent communication pathway.  An edge $e_{ji}=(v_j,v_i)$ means that agent $v_i$ is recipient of information from agent $v_j$. The set of direct predecessors of agent $v_i$ is denoted by 
$\mathcal{N}_{in}(v_i)=\{v_j |(v_j,v_i) \in\mathcal{E} \}$.

\noindent\textbf{Budget-Aware.} 
For query $\mathcal{Q}$, user has a willingness to pay or budget constraint $B$. The cost considered in this paper is monetary cost:
{\setlength{\abovedisplayskip}{3pt}\setlength{\belowdisplayskip}{3pt}\begin{equation}
C_i=c^{\mathrm{in}}_{m_i}T^{\mathrm{in}}_i+c^{\mathrm{out}}_{m_i}T^{\mathrm{out}}_i.
\end{equation}}%
Remaining budget $b_i$ is the money left before agent $i$, denoted as:
{\setlength{\abovedisplayskip}{3pt}\setlength{\belowdisplayskip}{3pt}\begin{equation}
b_i=B-\sum_{j<i}C_j.
\label{eq:cost}
\end{equation}}%

\noindent\textbf{Observations.} 
Executing agent $i$ therefore returns an observation. Given the byte counts of information $\bytes{\tilde y_i}$, denoted as:
{\setlength{\abovedisplayskip}{3pt}\setlength{\belowdisplayskip}{3pt}\begin{equation}
o_i=\Big(\ |\tilde y_i|_{\mathrm{B}}, T^{\mathrm{in}}_i,\ |y_i|_{\mathrm{chars}},\ T^{\mathrm{out}}_i\Big),
\end{equation}}%
where $T^{\mathrm{in}}_i$ is the token count of the message to $v_i$, $|\tilde y_i|_{\mathrm{B}}$ is the size in bytes of the message $v_i$ transmits to its successors, which would cost $v_i$'s successor to read, and agent $i$ outputs the message $y_i$, of length $|y_i|_{\mathrm{chars}}$ characters. $T^{\mathrm{out}}_i$ is the token count for output.  
A trajectory $\xi$ records the graph together with everything its execution produced: the
responses $y_i$, the LLM provider-reported token counts $T^{\mathrm{in}}_i$ and $T^{\mathrm{out}}_i$, and the per-call costs
\(
C(\xi)=\sum_{i=1}^{N}C_i.
\) 
\subsection{Budget-Aware Orchestration}
\noindent\textbf{Optimization Goal.}
Let $\mathcal{P}$ be the distribution over query--budget pairs. The ideal budget-aware MAS should be able to optimize a strategy to maximize utility by considering both performance and marginal cost under the given budget constraints:
{\setlength{\abovedisplayskip}{3pt}\setlength{\belowdisplayskip}{3pt}%
\begin{equation}
\begin{aligned}
\max_{\theta}\quad
&
\mathbb{E}_{(\mathcal{Q},B)\sim\mathcal{P}}
\mathbb{E}_{\xi\sim\pi_{\theta}(\cdot\mid \mathcal{Q},B)}
\left[U(\mathcal{Q},B,\xi)\right]
\\
\text{s.t.}\quad
&
\mathbb{E}_{\xi\sim\pi_{\theta}(\cdot\mid \mathcal{Q},B)}
\left[C(\mathcal{Q},\xi)\right]
\le B,
\quad
\forall\,(\mathcal{Q},B)\sim\mathcal{P},
\end{aligned}
\label{eq:objective}
\end{equation}}%
where $\pi_\theta$ is the policy distribution induced by the router and the execution environment.
The utility formalizes the trade-off between answer quality and monetary cost. We introduce the utility function:
{\setlength{\abovedisplayskip}{3pt}\setlength{\belowdisplayskip}{3pt}\begin{equation}
U(\mathcal{Q},B,\xi)=S(\mathcal{Q},\xi)-\gamma\frac{C(\xi)}{B},
\end{equation}}%
where $S(\mathcal{Q},\xi)\in\{0,1\}$ indicates whether the answer is correct
and $\gamma$ is tuning parameter reflecting the elasticity of cost sensitivity. The higher value of $\gamma$ penalizes expenditure more heavily.

\noindent\textbf{MAS Generation.}
In our work, rather than relying on fixed graph templates and deterministic model selection, \texttt{MeterRouter} constructs the MAS as an arbitrary DAG to consider a more general case.
To achieve this, \texttt{MeterRouter} incrementally expands the expert team by selecting each agent's role, backbone, and incoming edges under the remaining budget progressively.
To implement it, we learn a stochastic construction policy $\pi_\theta$. It induces a distribution over feasible orchestration graphs,
which is denoted by
$
P_\theta(\mathcal{G}\mid \mathcal{Q}, B, \mathcal{R}, \mathcal{M})=P_\theta(\mathcal{G}\mid \mathcal{S}).
$
We factorize this distribution into a tractable sequence of conditional decisions. At each step, the policy either terminates or adds a new agent with its role, backbone, and incoming edges, conditioned on the current state:
{\setlength{\abovedisplayskip}{3pt}\setlength{\belowdisplayskip}{3pt}  \begin{equation}
\small
P_\theta(\mathcal{G}\mid \mathcal{S})
=
\prod_{i\in\mathcal{I}}
\underbrace{
\pi_\theta(\tau_i \mid s_i)^{I_i^\tau}
}_{\text{Termination}}
\prod_{i=1}^{|\mathcal{V}|}
\left[
\underbrace{
\pi_\theta(r_i\mid s_i)
}_{\text{Role}}
\underbrace{
\pi_\theta(m_i\mid r_i,s_i)
}_{\text{Backbone}}
\prod_{j<i}
\underbrace{
\pi_\theta
\left(
e_{ji}\mid r_i,m_i,\mathbf{e}_{>j,i},s_i
\right)^{I_{ji}^e}
}_{\text{Edge}}
\right]
\label{eq:factorization}
\end{equation}}

Here, \(\mathcal I\) denotes the reached construction steps. The indicators \(I_i^\tau\) and \(I_{ji}^e\) exclude decisions deterministically imposed by structural or budget constraints from the policy probability. 
The state before generating the $i$-th agent, state:
{\setlength{\abovedisplayskip}{3pt}\setlength{\belowdisplayskip}{3pt}\begin{equation}
s_i =
\left(
\mathcal{Q},B,
b_i,
\mathcal{G}_{<i},
o_{<i}
\right),
\end{equation}}%
maintains accumulated knowledge and experience from previous interactions by containing the information of query $\mathcal{Q}$, remaining budget $b_i$, existing graph $\mathcal{G}_{<i}$, and previous steps observation $o_{<i}$.

There are three types of decisions in the construction:
\textbf{agent generation}, which predicts the role and selects a backbone
from the admissible set $\mathcal{M}_i^m\subseteq\mathcal{M}$ detailed in Section~\ref{sec:meter};
\textbf{edge generation}, which connects the incoming agent with existing
agents; and \textbf{termination} to stop construction.

\section{MeterRouter}
Based on the autoregressive graph generation paradigm, the graph generation is designed to implement the sequential generation process. \texttt{MeterRouter} instantiates $\pi_\theta$ as a small recurrent network (\S\ref{sec:arch}), making the agent aware of available resource limits (\S\ref{sec:meter}) by tracking the remaining budget.
It is trained in two stages and run in a generate--execute loop at inference (\S\ref{sec:train}). 
\subsection{Model Architecture}
\label{sec:arch}

\noindent\textbf{Input Representation.}
All textual conditioning information (query, role descriptions and the backbone) is mapped into dense vector representations. The query is mapped into a fixed-dimensional vector $\mathcal{Q} \to \mathbb{R}^D$ by a frozen sentence-BERT text encoder to extract the semantic information, followed by Layer Normalization (LN). $f_B$ denotes the $d$-dimensional embedding of the log budget, and $\tilde{f}_\mathcal{Q}$ fuses the query with the budget:
{\setlength{\abovedisplayskip}{3pt}\setlength{\belowdisplayskip}{3pt}\begin{equation}
f_\mathcal{Q}=\MLP\big(\LN(\Enc(\mathcal{Q}))\big), \quad
f_B=\MLP\big(\log B\big), \quad
\tilde{f}_\mathcal{Q}=\MLP\big([f_\mathcal{Q},f_B]\big).
\end{equation}}%

Similarly, each available agent role $r\in\mathcal{R}$ is converted into $z_r=\MLP\big(\LN(\Enc(r))\big)$, sharing $\MLP$ with the query. These are collected into a role embedding matrix $\mathbf{Z} \in \mathbb{R}^{|\mathcal{R}|\times d}$. For each LLM $m\in \mathcal{M}$, the model information is concatenated with its price,
{\setlength{\abovedisplayskip}{3pt}\setlength{\belowdisplayskip}{3pt}\begin{equation}
z_m=\MLP\big([\Enc(m),\log c^{\mathrm{in}}_m,\log c^{\mathrm{out}}_m]\big).
\end{equation}}%
Encoding the price into the latent vector represents each backbone by what it costs as well as by its description. The edge policy below additionally receives the receiver's input price explicitly.

\noindent\textbf{State Representation.}
A gated recurrent unit (GRU) aggregates the joint role--backbone encoding of the agents generated so far, $\{[z_{r_j},z_{m_j}]\}_{j<i}$, forming a historical embedding $f^{(i)}_{\mathrm{hist}}$ in Eq.~\ref{eq:hist}.
The meter enters as a cost vector $\phi_i$ holding the shares of the budget spent and remaining, and the shares of the agent cap used and remaining, formed as
{\setlength{\abovedisplayskip}{3pt}\setlength{\belowdisplayskip}{3pt}\begin{equation}
\phi_i=\Big[\frac{B-b_i}{B},\,\frac{b_i}{B},\,\frac{i-1}{N_{\max}},\;
\frac{N_{\max}-i+1}{N_{\max}},\;\psi(B)\Big]
\end{equation}}%
with $\psi(B)$ in~Eq.~\ref{eq:psi_b}, the log budget rescaled so that the training range $[B_{\min},B_{\max}]$ maps to $[-1,1]$.

$f^{(i)}_{\mathrm{cont}}$ combines the historical information with the current information by composing the budget-conditioned query through a dynamic gate, giving all context information and cost:
{\setlength{\abovedisplayskip}{3pt}\setlength{\belowdisplayskip}{3pt}\begin{equation}
f^{(i)}_{\mathrm{cont}}=(1-g_i)\cdot f^{(i)}_{\mathrm{hist}}+g_i\cdot\tilde{f}_\mathcal{Q}  +\MLP(\phi_i),
\quad
g_i=\sigma\!\left(\frac{f^{(i)}_{\mathrm{hist}}\cdot\tilde{f}_\mathcal{Q}}{\sqrt{d}}\right),
\end{equation}}%
where $\sigma$ represents the sigmoid function. $f^{(i)}_{\mathrm{edge}}$ in Eq.~\ref{eq:edge_emb} captures the connectivity pattern of the previous agent by collecting its incoming edges together with a mask of the predecessors already scanned, and encoding them using an MLP.

Another $\mathrm{GRU}$ then updates the node-level state $h^{(i)}_{\mathrm{node}}$ based on the context information together with the connectivity of the previous agent, denoted as
{\setlength{\abovedisplayskip}{3pt}\setlength{\belowdisplayskip}{3pt}\begin{equation*}
h^{(i)}_{\mathrm{node}}=\mathrm{GRU}\Big(\MLP\big([f^{(i)}_{\mathrm{cont}},\,f^{(i)}_{\mathrm{edge}}]\big),\,h^{(i-1)}_{\mathrm{node}}\Big).
\end{equation*}}%

\noindent\textbf{Role.}
A metric-learning module projects the hidden state $h^{(i)}_{\mathrm{node}}$ into a ``node intent'' embedding to predict a node prediction score $\omega^{i}_{\mathrm{node}}$. The $\operatorname{softmax}(\cdot)$ function converts the scores into the probability distribution for role selection:
{\setlength{\abovedisplayskip}{3pt}\setlength{\belowdisplayskip}{3pt}  \begin{equation}
\omega^{i}_{\mathrm{node}}=\frac{\MLP(h^{(i)}_{\mathrm{node}})^{\!\top}\MLP(z_r)}{\sqrt{d}},
\qquad
\pi_\theta(r_i =r \mid s_i) =\operatorname{softmax}_{r\in\mathcal{R}}\big(\omega^{i}_{\mathrm{node}}\big).
\label{eq:role}
\end{equation}}%

\noindent\textbf{Backbone.}
With the role fixed, the router assigns a backbone. Each candidate in the feasible set $\mathcal{M}^m_i$ is scored independently as $\omega_{m}^{i}$, so no role falls on an unaffordable backbone.
$\mathbf{l}_m = (\ell_m^{\mathrm{in}}, \ell_m^{\mathrm{out}})$ denotes the backbone's log price standardized across $\mathcal{M}$.
The pair $\boldsymbol{\psi}(B, b_i)= (\psi(B),\psi(b_i))$ reports the assigned budget and the realized remainder on one log-dollar scale. Because $b_i$ is computed from billed usage, it serves as execution feedback that enters the backbone decision:
{\setlength{\abovedisplayskip}{3pt}\setlength{\belowdisplayskip}{3pt}  \begin{equation}
\omega_{m}^{i}=\MLP\big([h^{(i)}_{\mathrm{node}},z_{r_i},z_m,f_B, \mathbf{l}_m
,\boldsymbol{\psi}(B, b_i)]\big),
\quad
\pi_\theta(m_i \mid r_i, s_i)=\operatorname{softmax}_{m\in\mathcal{M}^m_i}\Big(\omega_{m}^{i}
\Big).
\label{eq:route}
\end{equation}}%

\noindent\textbf{Edges.}
With $r_i$ and $m_i$ fixed, the router chooses which existing agents $v_{<i}$ to connect with. 
A GRU is employed to update the hidden state, initialized from the node-level state.
For existing agents $\{j=i-1,\dots,1\}$, scanned in this reverse order, the edge GRU at each step updates the state based on the previous edge decision and the predecessor's length vector:
{\setlength{\abovedisplayskip}{3pt}\setlength{\belowdisplayskip}{3pt}\begin{equation}
\begin{aligned}
h^{(i,i)}_{\mathrm{edge}}&=\MLP\big(\big[h^{(i)}_{\mathrm{node}},z_{r_i},z_{m_i}\big]\big),\\
h^{(i,j)}_{\mathrm{edge}}&=\mathrm{GRU}\Big(\MLP\big(\big[e_{j+1,i},z_{r_j},z_{m_j}\big]\big)+\MLP\big(\boldsymbol{\ell}_{j\to i}\big),\,h^{(i,j+1)}_{\mathrm{edge}}\Big).
\end{aligned}
\end{equation}}%
where a length vector $\boldsymbol{\ell}_{j\to i}$ is built from its observation $o_j$ and the receiver $m_i$ in Eq.~\ref{eq:length} to describe the resource usage of this edge.
The state $h^{(i,j)}_{\mathrm{edge}}$ then predicts the score for connecting to agent $j$, which gives the probability of edge $e_{ji}$:
{\setlength{\abovedisplayskip}{3pt}\setlength{\belowdisplayskip}{3pt}  \begin{equation}
\omega^{i,j}_{\mathrm{edge}}=\MLP(h^{(i,j)}_{\mathrm{edge}}),
\qquad
\pi_\theta\big(e_{ji}=1\mid r_i, m_i, \mathbf{e}_{>j,i}, s_i\big)=\sigma\big( \omega^{i,j}_{\mathrm{edge}}\big).
\end{equation}}%
When $v_i$ must close the graph, the edges from the current sinks are fixed to $1$ rather than sampled.

\noindent\textbf{Termination.}
\label{stop}
Team size is decided by the node state and the budget, so stopping can respond to a shrinking remainder in the cost vector $\phi_i$. The termination policy is evaluated at the start of step $i$.
Termination is forced when the affordable set $\mathcal{M}^m_i=\emptyset$.
The probability of termination is:
{\setlength{\abovedisplayskip}{3pt}\setlength{\belowdisplayskip}{3pt}  \begin{equation}
\pi_\theta(\tau_i=1\mid s_i)=\sigma\big(\MLP([h^{(i)}_{\mathrm{node}},f_B])\big).
\label{eq:stop}
\end{equation}}%
\noindent\textbf{Design Rationale.}
Since an agent's cost $C_i$ is known only after it is executed and the admissible set $\mathcal{M}^m_i$ in Eq.~\ref{eq:mask} depends on the remaining budget $b_i$, \texttt{MeterRouter} builds and executes agents one at a time and feeds $b_i$ to every decision head. Within each step, the decisions follow how cost arises: termination comes first because it depends only on the remaining budget, the role precedes the backbone because the capacity an agent needs depends on its role, and the backbone precedes the edges because reading a predecessor $v_j$ costs about $c^{\mathrm{in}}_{m_i}T^{\mathrm{out}}_j$, which the edge head sees through the length vector $\boldsymbol{\ell}_{j\to i}$ in Eq.~\ref{eq:length}. The budget enters through budget-relative shares and a log scale, and each role or backbone is scored by a shared function over the feasible set, so one model works across budgets and LLM pools of any size.
\subsection{The Meter}
\label{sec:meter}

Besides tracking the remaining budget in our policy condition, we confine generation to the affordable budget.
We therefore form budget awareness from two sides. One removes unaffordable backbones from the action set; the other converts the remaining money into a hard token cap on the call that is about to be issued. Both run on a byte-based estimator that is deliberately conservative; all constants and estimator forms in this section are given in Appendix~\ref{app:realize}.

\noindent\textbf{Affordable backbone set.}
Let $\widehat{T}_i$ be an upper estimation of the input node $i$ would consume, if it gets all messages from every completed predecessor. 
Let $F_m(T)=c^{\mathrm{in}}_mT+c^{\mathrm{out}}_mq_{\min}$ be the floor cost of a call on $m$
that produces the predefined minimum admissible completion constant $q_{\min}$. Then:
{\setlength{\abovedisplayskip}{3pt}\setlength{\belowdisplayskip}{3pt}  \begin{equation}
\mathcal{M}^m_i=\Big\{m\in\mathcal{M}\;:\;F_m\big(\widehat{T}_i\big)+R_i\le b_i\Big\},
\label{eq:mask}
\end{equation}}%
where $R_i$ reserves the floor cost of any successor. If $\mathcal{M}^m_i=\emptyset$, the step terminates. If there is no affordable LLM before the first agent, the query is flagged as budget-blocked without issuing any model call. %

\noindent\textbf{Priced completion cap.}
Once the role, the backbone and the edges are fixed, the input message is decided and the remaining budget becomes the hard cap on completion. So the cap decides how long the node $i$ may output in the remaining budget:
{\setlength{\abovedisplayskip}{3pt}\setlength{\belowdisplayskip}{3pt}  \begin{equation}
q_i=\min\Big\{T_{\max},\;\Big\lfloor\frac{b_i-\hat\rho_i-R_i}{\mu_i}\Big\rfloor\Big\},
\qquad
\mu_i=c^{\mathrm{out}}_{m_i}+\nu_i\,\kappa_{\mathrm{r}}\,c^{\mathrm{in}}_{\underline{m}},
\label{eq:cap}
\end{equation}}%
where  $T_{\max}$ is the maximum number of completion tokens allowed per call, $\hat\rho_i$ is the estimated input charge, $\nu_i= 1$ if a successor is still obliged, $\underline{m}$ is the cheapest backbone and $\kappa_{\mathrm{r}}$ converts output tokens into the input tokens they become downstream. The marginal price $\mu_i$ can help tighten the cap on exactly the agents whose output will be re-read. Under the estimator, a shielded trajectory satisfies $C(\xi)\le B$ by construction. The estimator is an upper bound.
\subsection{Policy Procedure}
\label{sec:train}

\texttt{MeterRouter} is trained in two separate stages to balance a good initialization against exploration of the full action space. In the \textbf{first stage}, the policy learns a prior MAS distribution from a set of pre-generated supervision MAS, $\mathcal{D}_{\mathrm{SFT}}$, by a weighted negative log-likelihood objective function on role and edge decisions in Eq.~\ref{eq:sft} inspired by \cite{argdesigner} as shown in Algorithm~\ref{alg:stage1}. This loss has no budget or backbone term, so the budget enters only by the label in $\mathcal{D}_{\mathrm{SFT}}$. 

To \textbf{build} $\mathbf{\mathcal{D}_{\mathrm{SFT}}}$, for every query $\mathcal{Q}$, each template $t$ in a predefined pool $\mathcal{T}$ listed in Table~\ref{tab:graph_templates} is implemented $K$ times. Each run draws the roles uniformly and every agent is on the mid-priced backbone. The runs are executed once and re-scored at every budget $B$ of a fixed log-spaced grid by the empirical utility, 
{\setlength{\abovedisplayskip}{3pt}\setlength{\belowdisplayskip}{3pt}\begin{equation}
\begin{gathered}
\hat u_t(\mathcal{Q},B)=\frac{1}{K}\sum_{k=1}^{K}S(\mathcal{Q},\xi_{t,k})-\frac{\gamma}{B}\cdot\frac{1}{K}\sum_{k=1}^{K}C(\xi_{t,k}),\\
t^\star(\mathcal{Q},B)=\argmax_{t\in\mathcal{T}:\ \hat c_t\le B}\ \hat u_t(\mathcal{Q},B),\qquad \hat c_t=\frac{1}{K}\sum_{k=1}^{K}C(\xi_{t,k}).
\end{gathered}
\end{equation}}%
We will always put the templates that maximize $\hat u_t(\mathcal{Q},B)$ and use cost less than $B$ into $\mathcal{D}_{\mathrm{SFT}}$, as detailed in Algorithm~\ref{alg:dsft}. Because the label changes with $B$, $\mathcal{D}_{\mathrm{SFT}}$ pairs every query with its highest-utility affordable template at each budget. The backbone policy is frozen in this stage, so supervision shapes topology and roles, and backbone allocation is left to the second stage.

In the \textbf{second stage}, the policy is refined by on-policy Reinforcement Learning (RL) since its answer can be graded quickly, so that backbone allocation and the response to execution feedback can be learned. For every training query $\mathcal{Q}$, a budget $B$ is drawn log-uniformly from $[B_{\min},B_{\max}]$ to get a better ability of generalization. The current policy $\pi_{\theta_{\mathrm{old}}}$ then generates and executes a group of $G$ trajectories $\{\xi_g\}_{g=1}^G$ under this common budget. Each trajectory is scored by the utility $U_g=U(\mathcal{Q},B,\xi_g)$, and its advantage in its own group is denoted as:
\(
A_g=U_g-\frac{1}{G}\sum_{g'=1}^{G}U_{g'}.
\)%
Let $\mathcal{K}(\xi)$ collect the decisions of $\xi$ made from the policy and $\rho_d(\theta)=\pi_\theta(d)/\pi_{\theta_{\mathrm{old}}}(d)$ be the ratio of $\pi_\theta$ and $\pi_{\theta_{\mathrm{old}}}$ assigned to decision $d$. The policy is updated with the clipped surrogate:
{\setlength{\abovedisplayskip}{3pt}\setlength{\belowdisplayskip}{3pt}  \begin{equation}
\mathcal{L}_{\mathrm{RL}}=-\,\mathbb{E}_{\mathcal{Q},B,\,\xi_{1:G}\sim\pi_{\theta_{\mathrm{old}}}}\Bigg[\frac{1}{G}\sum_{g=1}^{G}\frac{1}{Z}\sum_{d\in\mathcal{K}(\xi_g)}\min\Big\{\rho_d(\theta)A_g,\ \operatorname{clip}\big(\rho_d(\theta),1-\epsilon,1+\epsilon\big)A_g\Big\}\Bigg],
\label{eq:rl}
\end{equation}}%
where $Z$ is the mean number of decisions in the demonstrations of the first stage, a constant fixed once. The objective function follows Dr.~GRPO~\citep{drgrpo} to reward the improvement of utility and isolate the LLM providers' contribution to the likelihood of $\xi$, which does not depend on $\theta$ and is totally black box. By considering the ratios $\rho_d$, we therefore obtain the exact likelihood ratio of the whole trajectory. All parameters are trained in this stage. The detail can be seen in Algorithm~\ref{alg:stage2}.

At \textbf{inference} the parameters are frozen, and the router receives only the query $\mathcal{Q}$ and the budget $B$. It builds the multi-agent system online by sampling from $\pi_\theta$, which makes the decision of every step based on the MAS structure and all responses before. As each agent is executed and metered before the next decision is made, every later decision can be made on what the earlier calls actually cost and returned. The answer is the output of the final agent. \texttt{MeterRouter} takes the budget $B$ as an input, so sweeping $B$ yields a set of MAS along a cost–accuracy curve from one trained model. Detailed information of it is in Algorithm~\ref{alg:inference}. 

\section{Experiments}
\subsection{Experiments Setup}
\label{sec:exp_setup}

\noindent\textbf{Datasets.}
We evaluate \texttt{MeterRouter} on three types of tasks:
(1) general reasoning, with MMLU-Pro~\citep{mmlupro},
MuSR~\citep{musr}, and BBEH~\citep{bbeh};
(2) mathematical reasoning, with Omni-MATH-2~\citep{omnimath2};
and (3) code generation, with MBPP+~\citep{evalplus}.
For each benchmark, we split the queries into disjoint training,
validation, and test sets (Table~\ref{tab:dataset-split}).
Omni-MATH-2, MMLU-Pro, and BBEH each have 1,000 test queries,
and MuSR and MBPP+ use every query left after the training and
validation splits as their test sets (516 and 138).

\noindent\textbf{Baselines.}
We compare \texttt{MeterRouter} with four groups of methods:
(1) single-agent methods, namely vanilla and Chain-of-Thought
prompting~\citep{cot};
(2) fixed multi-agent topologies, including Chain, Tree, and Complete
Graph~\citep{macnet}, and LLM-Debate~\citep{llmdebate};
(3) dynamic MAS, including AutoGen~\citep{autogen},
GPTSwarm~\citep{gptswarm}, and AgentPrune~\citep{agentprune};
and (4) LLM routing methods, including GraphRouter~\citep{graphrouter},
RouterDC~\citep{routerdc}, MasRouter~\citep{masrouter}, and BAMAS~\citep{bamas}, which explicitly includes the budget.

\noindent\textbf{Backbone LLMs.}
We build a three-tier LLM pool from gpt-oss-20b, gpt-oss-120b, and MiniMax-M3 (Table~\ref{tab:model-cards}). Their sizes grow from 20B total parameters for gpt-oss-20b to $\sim$428B for MiniMax-M3. The two gpt-oss models support 128K-token contexts, while MiniMax-M3 supports up to 1M tokens. Per-token prices rise with tier: from gpt-oss-20b to MiniMax-M3, the input price increases from \$0.06 to \$0.30 and the output price from \$0.19 to \$1.20 per million tokens. The model support an extensible LLM pool (Appendix ~\ref{app:pool}).

\noindent\textbf{Implementation.} The text encoder is all-MiniLM-L6-v2~\citep{wang2020minilm} and each system has at most $N_{\max}=5$ agents. Role sets follow the style of ARG-Designer~\citep{argdesigner}. All LLMs are called with temperature $0.2$  and at most $T_{\max}=32{,}768$ completion tokens per call. We evaluate at $B\in\{1,5,50\}\times10^{-3}$ and add $B=4\times10^{-3}$ for Omni-MATH-2 and BBEH, where accuracy changes most sharply between $1$ and $5\times10^{-3}$. Each benchmark's checkpoint is selected on the validation set, and we report results on the test sets. We access all LLMs through Ollama Cloud API.

\subsection{Performance Analysis}

\begin{table*}[ht]
\centering
\caption{Comparison with single-agent, multi-agent, and routing baselines.
\textbf{Multi.}: builds a multi-agent topology;
\textbf{Router}: selects LLMs from the pool.
Acc.\ is averaged over all test questions, and Cost is the average cost per query. The top five performing methods are highlighted in bold.}
\label{tab:baseline-results}

\renewcommand{\arraystretch}{1.25}

\resizebox{\textwidth}{!}{%
\begin{tabular}{lcclccccccc}
\toprule

\rowcolor{gray!35}
\textbf{Method}
& \textbf{Multi.}
& \textbf{Router}
& \textbf{LLM}
& \textbf{Omni-Math-2}
& \textbf{MMLU-Pro}
& \textbf{MuSR}
& \textbf{BBEH}
& \textbf{MBPP+}
& \textbf{Acc.}
& \textbf{Cost (\$ $\times 10^{-3}$)} \\

\midrule

\rowcolor{gray!12}
& & & MiniMax-M3
& 43.1 & 80.6 & 63.2 & 27.8 & 79.7 & 53.39 & 4.14 \\

\rowcolor{gray!12}
& & & gpt-oss-120b
& 60.3 & 79.1 & 68.7 & 42.7 & 76.5 & 62.50 & 1.91 \\

\rowcolor{gray!12}
\multirow{-3}{*}{Vanilla}
& \multirow{-3}{*}{\xmk}
& \multirow{-3}{*}{\xmk}
& gpt-oss-20b
& 43.6 & 64.4 & 58.5 & 16.0 & 79.0 & 45.18 & 0.14 \\

& & & MiniMax-M3
& 39.5 & 80.4 & \textbf{70.0} & 29.9 & 76.8 & 53.78 & 2.61 \\

& & & gpt-oss-120b
& 61.8 & 78.3 & \textbf{71.5} & 42.4 & 78.0 & 62.98 & 1.84 \\

\multirow{-3}{*}{CoT}
& \multirow{-3}{*}{\xmk}
& \multirow{-3}{*}{\xmk}
& gpt-oss-20b
& 44.2 & 68.1 & 61.8 & 18.1 & 76.8 & 47.32 & 0.17 \\

\midrule

\rowcolor{gray!12}
& & & MiniMax-M3
& 58.3 & \textbf{83.3} & 66.3 & 40.7 & 81.2 & 62.32 & 12.68 \\

\rowcolor{gray!12}
& & & gpt-oss-120b
& 57.7 & 78.2 & 63.2 & 36.8 & 81.9 & 59.28 & 3.65 \\

\rowcolor{gray!12}
\multirow{-3}{*}{Chain}
& \multirow{-3}{*}{\cmk}
& \multirow{-3}{*}{\xmk}
& gpt-oss-20b
& 51.3 & 70.5 & 64.0 & 23.2 & 81.9 & 51.81 & 1.13 \\

& & & MiniMax-M3
& 59.8 & \textbf{83.7} & \textbf{69.8} & 41.2 & 82.6 & 63.52 & 16.87 \\

& & & gpt-oss-120b
& 57.3 & 78.2 & 65.9 & 36.7 & \textbf{84.1} & 59.61 & 4.30 \\

\multirow{-3}{*}{Tree}
& \multirow{-3}{*}{\cmk}
& \multirow{-3}{*}{\xmk}
& gpt-oss-20b
& 50.9 & 71.1 & 60.3 & 23.6 & 81.2 & 51.42 & 1.38 \\

\rowcolor{gray!12}
& & & MiniMax-M3
& 54.5 & \textbf{83.0} & 66.1 & 35.4 & \textbf{84.8} & 59.85 & 13.03 \\

\rowcolor{gray!12}
& & & gpt-oss-120b
& 55.6 & 78.1 & 63.8 & 34.7 & 82.6 & 58.21 & 3.24 \\

\rowcolor{gray!12}
\multirow{-3}{*}{Complete Graph}
& \multirow{-3}{*}{\cmk}
& \multirow{-3}{*}{\xmk}
& gpt-oss-20b
& 49.6 & 69.5 & 62.2 & 23.2 & 80.4 & 50.77 & 1.01 \\

& & & MiniMax-M3
& 60.6 & \textbf{82.6} & 62.8 & 38.2 & 83.3 & 61.65 & 27.34 \\

& & & gpt-oss-120b
& \textbf{65.7}
& 80.0
& \textbf{69.5}
& \textbf{51.3}
& 78.7
& 66.70
& 10.13 \\

\multirow{-3}{*}{LLM-Debate}
& \multirow{-3}{*}{\cmk}
& \multirow{-3}{*}{\xmk}
& gpt-oss-20b
& 53.0 & 70.8 & 60.7 & 20.6 & 80.4 & 51.12 & 0.90 \\

\midrule

\rowcolor{gray!12}
& & & MiniMax-M3
& 42.4 & 80.8 & 63.6 & 29.9 & 79.0 & 53.86 & 6.10 \\

\rowcolor{gray!12}
& & & gpt-oss-120b
& 63.1 & 79.0 & 66.7 & 46.6 & 82.6 & 64.18 & 3.79 \\

\rowcolor{gray!12}
\multirow{-3}{*}{AutoGen}
& \multirow{-3}{*}{\cmk}
& \multirow{-3}{*}{\xmk}
& gpt-oss-20b
& 47.6 & 68.0 & 58.3 & 18.3 & 81.2 & 47.95 & 0.36 \\

& & & MiniMax-M3
& 45.9 & 81.9 & 67.6 & 36.1 & \textbf{84.0} & 57.57 & 6.93 \\

& & & gpt-oss-120b
& \textbf{63.9}
& 78.5
& 68.3
& \textbf{47.0}
& 81.6
& 64.56
& 6.74 \\

\multirow{-3}{*}{GPTSwarm}
& \multirow{-3}{*}{\cmk}
& \multirow{-3}{*}{\xmk}
& gpt-oss-20b
& 48.0 & 70.8 & 62.0 & 17.3 & 77.5 & 48.93 & 0.52 \\

\rowcolor{gray!12}
& & & MiniMax-M3
& 47.1 & 80.8 & 66.5 & 30.4 & 81.9 & 55.80 & 10.87 \\

\rowcolor{gray!12}
& & & gpt-oss-120b
& \textbf{65.9}
& 79.6
& 68.4
& \textbf{48.3}
& \textbf{84.8}
& 65.90
& 8.33 \\

\rowcolor{gray!12}
\multirow{-3}{*}{AgentPrune}
& \multirow{-3}{*}{\cmk}
& \multirow{-3}{*}{\xmk}
& gpt-oss-20b
& 48.6 & 71.0 & 57.2 & 18.8 & 77.5 & 48.88 & 0.72 \\

\midrule

\multirow{1}{*}{GraphRouter}
& \multirow{1}{*}{\xmk}
& \multirow{1}{*}{\cmk}
& LLM Pool
& 44.2 & 65.2 & 58.7 & 24.7 & 78.9 & 47.97 & 0.35 \\

\rowcolor{gray!12}
\multirow[t]{1}{*}{RouterDC}
& \multirow[t]{1}{*}{\xmk}
& \multirow[t]{1}{*}{\cmk}
& LLM Pool
& 44.0 & 72.6 & 60.7 & 25.3 & 81.3 & 50.39 & 0.24 \\

\multirow{1}{*}{BAMAS}
& \multirow{1}{*}{\cmk}
& \multirow{1}{*}{\cmk}
& LLM Pool
& 45.3 & 78.7 & 67.2 & 35.9 & 76.8 & 56.15 & 4.93 \\

\rowcolor{gray!12}
\multirow{1}{*}{MasRouter}
& \multirow{1}{*}{\cmk}
& \multirow{1}{*}{\cmk}
& LLM Pool
& \textbf{67.4}
& 80.8
& 67.8
& \textbf{49.3}
& 82.6
& 66.80
& 14.15 \\

\midrule

\multirow{1}{*}{\textbf{MeterRouter}}
& \multirow{1}{*}{\cmk}
& \multirow{1}{*}{\cmk}
& LLM Pool
& \textbf{64.2}
& \textbf{83.6}
& \textbf{72.2}
& \textbf{46.8}
& \textbf{86.6}
& 66.72
& 2.56 \\

\bottomrule
\end{tabular}%
}
\end{table*}

\noindent\textbf{Accuracy.} Table~\ref{tab:baseline-results} reports test accuracy and cost per query, with \texttt{MeterRouter} at $B=5\times10^{-2}$ and $N_{\max}=5$. \texttt{MeterRouter} reaches an average accuracy of 66.72\%, 0.08 points below the most accurate baseline, MasRouter (66.80\%), and 0.02 points above LLM-Debate with gpt-oss-120b (66.70\%). Baseline results with three seeds in Appendix~\ref{app:seed}. It is the most accurate method on MuSR (72.2\%) and MBPP+ (86.6\%). BBEH is its weakest benchmark: there, LLM-Debate with gpt-oss-120b is 4.5 points more accurate.

\begin{figure*}[ht]
    \centering
    \includegraphics[width=0.70\textwidth]{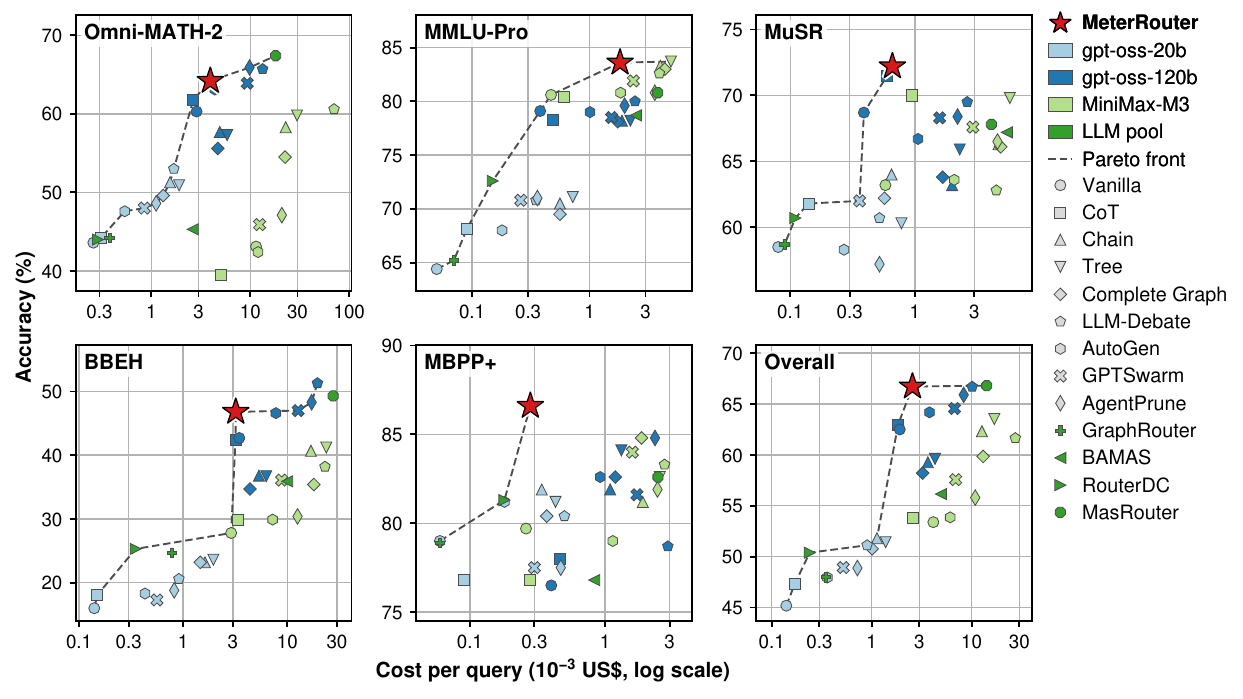}
    \caption{Accuracy versus cost per query on each benchmark and on the overall average. The dashed line is the Pareto frontier; marker shape denotes the method and color the backbone LLM.}
    \label{fig:cost_accuracy}
\end{figure*}

\noindent\textbf{Cost--accuracy trade-off.} \texttt{MeterRouter} spends $2.56\times10^{-3}$ \$ per query on average, $4\times$ less than LLM-Debate and $5.5\times$ less than MasRouter. Fig.~\ref{fig:cost_accuracy} shows that it lies on the Pareto frontier of all five benchmarks and of the overall average, so no baseline is both more accurate and cheaper. On the overall front, \texttt{MeterRouter} sits at the point where extra spending stops paying off: LLM-Debate with gpt-oss-120b costs about $4\times$ more but is 0.02 points less accurate. Compared with BAMAS, whose predefined tier-based preference systematically favors high-tier models (MiniMax-M3), our method can instead select mid- or low-tier models when they perform better.  

\subsection{Ablation Study}
{\setlength{\intextsep}{2pt}%
\begin{table}[ht]
\centering
\caption{Ablation studies of \texttt{MeterRouter}. Cost is measured in units of \$ $\times 10^{-3}$.}
\label{tab:ablation}
\setlength{\aboverulesep}{1pt}
\setlength{\belowrulesep}{1pt}
\resizebox{\linewidth}{!}{%
\begin{tabular}{lccccccc}
\toprule
\textbf{Method}
& \textbf{OmniMath-2}
& \textbf{MMLU-Pro}
& \textbf{MuSR}
& \textbf{BBEH}
& \textbf{MBPP+}
& \textbf{Acc.}
& \textbf{Cost} \\
\midrule

MeterRouter
& 64.2 & 83.6 & 72.2 & 46.8 & 86.6 & 66.72 & 2.56 \\
\midrule

w/o RL
& 53.7 & 75.9 & 67.2 & 34.2 & 79.2 & 57.34 & 2.55 \\

$B=\$5\times10^{-3}$
& 60.4 & 81.4 & 69.4 & 35.0 & 85.5 & 61.42 & 1.96 \\

$B\in\{\$1,\$4\}\times10^{-3}$
& 56.2 & 73.8 & 62.4 & 31.3 & 82.2 & 56.06 & 1.53 \\

\midrule

Max agents = 4
& 65.5 & 82.2 & 71.6 & 45.2 & 84.8 & 66.08 & 2.47 \\

\quad w/o RL
& 52.1 & 74.5 & 64.0 & 33.5 & 81.2 & 55.92 & 2.84 \\

\quad $B=\$5\times10^{-3}$
& 58.0 & 78.0 & 68.8 & 33.7 & 83.3 & 59.30 & 1.47 \\

\quad $B\in\{\$1,\$4\}\times10^{-3}$
& 56.3 & 73.4 & 58.7 & 32.5 & 80.8 & 55.73 & 1.27 \\

\midrule

Max agents = 6
& 65.1 & 83.1 & 69.1 & 45.1 & 84.2 & 65.85 & 3.44 \\

\quad w/o RL
& 53.8 & 75.0 & 65.7 & 34.7 & 79.9 & 57.07 & 3.19 \\

\quad $B=\$5\times10^{-3}$
& 57.4 & 79.9 & 66.3 & 35.8 & 83.2 & 59.90 & 1.81 \\

\quad $B\in\{\$1,\$4\}\times10^{-3}$
& 54.3 & 76.3 & 63.6 & 31.6 & 82.1 & 56.49 & 1.64 \\

\bottomrule
\end{tabular}%
}
\end{table}}

Table~\ref{tab:ablation} varies one factor at a time: whether RL is used, the budget, and the maximum number of agents $N_{\max}$. SFT learns a prior over topologies and roles but leaves model routing at its uniform initialization; RL then fine-tunes the whole policy, learning the LLM allocation and how to respond to the realized cost. At $N_{\max}=5$, replacing RL with random LLM selection (w/o RL) lowers average accuracy from 66.72\% to 57.34\%, with drops on all five benchmarks, while cost barely changes ($2.55$ vs.\ $2.56\times10^{-3}$ \$ per query). RL gives a similar gain at the other two agent caps. %

We run two further ablations to separate the two stages. First, to separate graph construction from LLM allocation, we train \texttt{MeterRouter} with a single backbone (gpt-oss-120b), so it builds the graph but cannot choose models (Appendix~\ref{app:sft_single}). It is only 3.19 points below the full model, and it is more accurate and cheaper than the same-backbone Chain, Tree, and Complete Graph baselines. Second, we fix the topology and train only the LLM allocation with RL (Appendix~\ref{app:fixed_topo}). This variant lies on the cost--accuracy frontier in Fig.~\ref{fig:fixed_topo}, so RL alone learns an efficient allocation. However, fixing the topology costs 7.03 to 7.51 points of accuracy, more than twice the drop from using a single backbone. Graph construction therefore accounts for most of the gain, and LLM allocation adds the rest.

\noindent\textbf{Budget compliance.} Nearly every query stays within its budget. The in-budget rate is at least 99.98\% at every tested budget (Table~\ref{tab:stop}). The rare overruns occur because one byte-based estimate is used for all LLMs, while each LLM counts input tokens differently. Counting input tokens with each LLM's own tokenizer before the call would make the estimate exact without changing the framework. Every call would then satisfy $C_i\le c^{\mathrm{in}}_{m_i}T^{\mathrm{in}}_i+c^{\mathrm{out}}_{m_i}q_i\le b_i$, so no query could exceed its budget. Appendix~\ref{app:stop} reports how often each query ends by a learned stop, the agent cap, or the budget, and how often answers are truncated or exceed the budget.

\begin{figure*}[ht]
\centering

\begin{minipage}[t]{0.55\textwidth}
    \centering
    \setlength{\abovecaptionskip}{2pt} 
    \includegraphics[width=\linewidth]{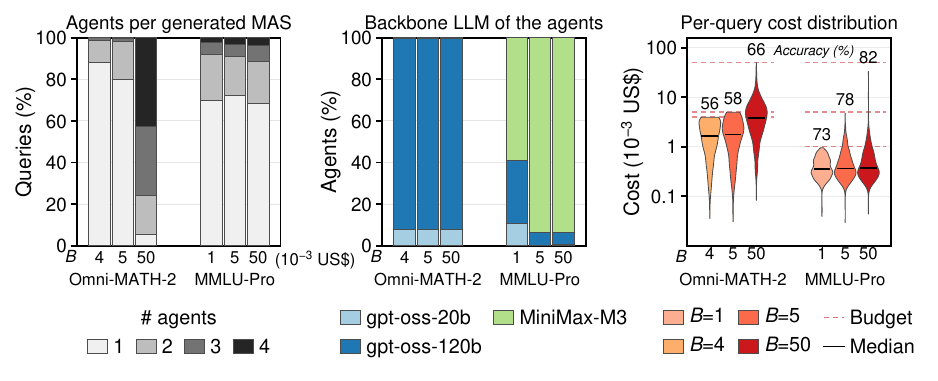}
    \caption{How \texttt{MeterRouter} adapts the MAS it generates to the per-query budget $B$ on OmniMath-2 and MMLU-Pro. \textbf{Left}: share of queries by the number of agents used. \textbf{Middle}: share of agents assigned to each backbone LLM. \textbf{Right}: distribution of per-query cost on a log scale. Each violin spans the min to max cost.}
    \label{fig:agent}
\end{minipage}
\hfill
\begin{minipage}[t]{0.43\textwidth}
    \centering
    \setlength{\abovecaptionskip}{2pt} 
    \includegraphics[width=\linewidth]{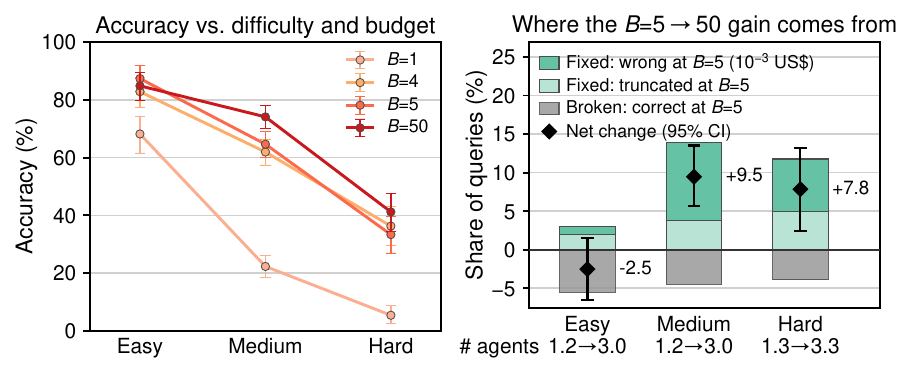}
    \caption{Budget use across problem difficulty on OmniMath-2 (Easy $<4$, Medium $4$--$5.5$, Hard $\geq5.5$). \textbf{Left}: accuracy of \texttt{MeterRouter} at each budget. \textbf{Right}: change from ${B=5\times10^{-3}}$ to ${B=5\times10^{-2}}$.}
    \label{fig:difficulty}
\end{minipage}

\end{figure*}

\noindent\textbf{Agent cap.} The agent cap has little impact on accuracy because it is rarely reached. As shown in Fig.~\ref{fig:agent} for MMLU-Pro, most runs terminate through the learned stopping policy before hitting the maximum number of agents. Across $N_{\max}\in\{4,5,6\}$, average accuracy varies by less than one point (65.85\% to 66.72\%, highest at $N_{\max}=5$), so the policy sets the graph size itself and does not depend on a tuned cap.

\noindent\textbf{Budget allocation.} Fig.~\ref{fig:agent} examines how \texttt{MeterRouter} allocates additional budget on Omni-MATH-2 and MMLU-Pro when $N_{\max}=4$, the two benchmarks whose generated systems change most with the budget. The two benchmarks spend the extra budget in different ways. On Omni-MATH-2, the router keeps gpt-oss-120b for about 92\% of agents at every budget but builds larger systems: the mean number of agents rises from 1.22 at $B=5\times10^{-3}$ to 3.13 at $5\times10^{-2}$, and 42.5\% of queries use four agents. On MMLU-Pro, the team size stays near 1.4 agents, and the budget instead buys a stronger backbone: the share of agents on MiniMax-M3 rises from 59\% at $B=10^{-3}$ to 93\% at $5\times10^{-3}$. The right panel shows that cost stays within the budget at every level, but its distribution changes with the budget. 
spending clusters near $B$ under tight budgets but varies more across queries at $B=5\times10^{-2}$, where the budget is no longer binding.

\noindent\textbf{Gains by difficulty.}
Omni-MATH-2 assigns each query a difficulty level, and Fig.~\ref{fig:difficulty} breaks down \texttt{MeterRouter}'s accuracy by difficulty and budget. The left panel shows that accuracy falls with difficulty at every budget and that the budget matters more on harder problems. The right panel shows that the larger budget, together with larger agent teams, improves accuracy on medium and hard problems but gives little benefit on easy ones. 

\noindent\textbf{Changing prices and extensible LLM pools.} \texttt{MeterRouter} handles new LLM prices and new LLMs without any change to its architecture. On MMLU-Pro, after price-paired training, the share of queries that assign gpt-oss-120b to the first agent falls from 92.1\% to 0.0\% as its price rises from $0.1\times$ to $10\times$ (Appendix~\ref{app:price}). Retrained with a fourth backbone, Qwen3.8-27B, the router gains 2.1 points on Omni-MATH-2 and 2.5 points on BBEH at 8\% and 2\% higher cost, and it stays on the Pareto frontier (Appendix~\ref{app:pool}).

\section{Conclusion}
We presented \texttt{MeterRouter}, which builds a multi-agent system for each query under a monetary budget, choosing each agent's role, backbone LLM, and incoming connections one at a time. Because every call is metered before the next decision, the router spends the remaining budget based on actual costs, and overruns are rare. Across five benchmarks, it matches the strongest multi-agent baselines in average accuracy at a fraction of their cost and lies on the cost--accuracy Pareto frontier.

\clearpage

\section*{AI Use Statement}
LLMs play two distinct roles in this work.
\emph{As objects of study:} gpt-oss-20b, gpt-oss-120b, and MiniMax-M3
serve as agent backbones in MeterRouter and in all baselines. They are
accessed through the Ollama Cloud API with the settings in
Section~\ref{sec:exp_setup}, and all reported accuracies and costs are produced by
these calls.
\emph{As writing assistants:} ChatGPT and Claude were used to
polish the grammar and wording of author-written text.

\section*{Ethics Statement}
This work aims to improve the cost efficiency and accessibility of multi-agent language model systems through budget-aware routing. Our experiments use publicly available benchmarks and focus on routing among frozen language models. We do not anticipate specific ethical concerns arising from the proposed routing method. However, MeterRouter may inherit biases, factual errors, and harmful outputs from its underlying models. Responsible deployment should therefore include appropriate safeguards, privacy protections for queries sent to external providers, and human oversight in sensitive applications

\section*{Reproducibility Statement}
We have made efforts to ensure the reproducibility of our results.
The full method, including the router architecture, the budget meter,
and both training stages, is specified in Section~4, with supplementary functions in Appendix~\ref{app:realize} and pseudocode in Appendix~\ref{app:alg}. All five benchmarks are publicly available; data splits are given in Table~\ref{tab:dataset-split}
and the complete role prompts in Appendix~\ref{app:role}. All backbone LLMs are accessed through Ollama Cloud API, and costs are computed at the fixed per-token rates in Table~\ref{tab:model-cards}. Hyperparameters of MeterRouter are listed in Table~\ref{tab:estimator}, and the configurations of all baselines in Appendix~\ref{app:baseline}. Source code will be made publicly available.

\bibliographystyle{iclr2027_conference}

\begingroup
\small
\setlength{\bibsep}{1pt plus 0.3ex}
\bibliography{iclr2027_conference}

@inproceedings{agentprune,
  author    = {Guibin Zhang and Yanwei Yue and Zhixun Li and Sukwon Yun and
               Guancheng Wan and Kun Wang and Dawei Cheng and Jeffrey Xu Yu and
               Tianlong Chen},
  title     = {Cut the Crap: An Economical Communication Pipeline for {LLM}-based
               Multi-Agent Systems},
  booktitle = {International Conference on Learning Representations ({ICLR})},
  publisher = {OpenReview.net},
  year      = {2025},
  url       = {https://openreview.net/forum?id=LkzuPorQ5L}
}

@inproceedings{agentdropout,
  author    = {Zhexuan Wang and Yutong Wang and Xuebo Liu and Liang Ding and
               Miao Zhang and Jie Liu and Min Zhang},
  title     = {{AgentDropout}: Dynamic Agent Elimination for Token-Efficient and
               High-Performance {LLM}-Based Multi-Agent Collaboration},
  booktitle = {Proceedings of the 63rd Annual Meeting of the Association for
               Computational Linguistics (Volume 1: Long Papers)},
  year      = {2025},
  publisher = {Association for Computational Linguistics},
  url       = {https://aclanthology.org/2025.acl-long.1170/},
  doi       = {10.18653/v1/2025.acl-long.1170},
  pages     = {24013--24035}
}

@inproceedings{gptswarm,
  title     = {{GPTSwarm}: Language Agents as Optimizable Graphs},
  author    = {Zhuge, Mingchen and Wang, Wenyi and Kirsch, Louis and Faccio, Francesco
               and Khizbullin, Dmitrii and Schmidhuber, J{\"u}rgen},
  booktitle = {Proceedings of the 41st International Conference on Machine Learning},
  pages     = {62743--62767},
  year      = {2024},
  volume    = {235},
  series    = {Proceedings of Machine Learning Research},
  publisher = {PMLR},
  url       = {https://proceedings.mlr.press/v235/zhuge24a.html}
}

@inproceedings{gdesigner,
  title     = {{G-Designer}: Architecting Multi-agent Communication Topologies via
               Graph Neural Networks},
  author    = {Zhang, Guibin and Yue, Yanwei and Sun, Xiangguo and Wan, Guancheng and
               Yu, Miao and Fang, Junfeng and Wang, Kun and Chen, Tianlong and
               Cheng, Dawei},
  booktitle = {Proceedings of the 42nd International Conference on Machine Learning},
  pages     = {76678--76692},
  year      = {2025},
  volume    = {267},
  series    = {Proceedings of Machine Learning Research},
  publisher = {PMLR},
  url       = {https://proceedings.mlr.press/v267/zhang25cu.html}
}

@inproceedings{argdesigner,
  author    = {Li, Shiyuan and Liu, Yixin and Wen, Qingsong and Zhang, Chengqi and
               Pan, Shirui},
  title     = {Assemble Your Crew: Automatic Multi-agent Communication Topology
               Design via Autoregressive Graph Generation},
  booktitle = {Proceedings of the AAAI Conference on Artificial Intelligence},
  volume    = {40},
  number    = {28},
  pages     = {23142--23150},
  year      = {2026},
  doi       = {10.1609/aaai.v40i28.39481}
}

@inproceedings{hybridllm,
  author    = {Dujian Ding and Ankur Mallick and Chi Wang and Robert Sim and
               Subhabrata Mukherjee and Victor R{\"u}hle and Laks V. S. Lakshmanan
               and Ahmed Hassan Awadallah},
  title     = {Hybrid {LLM}: Cost-Efficient and Quality-Aware Query Routing},
  booktitle = {International Conference on Learning Representations ({ICLR})},
  publisher = {OpenReview.net},
  year      = {2024},
  url       = {https://openreview.net/forum?id=02f3mUtqnM}
}

@inproceedings{routellm,
  author    = {Isaac Ong and Amjad Almahairi and Vincent Wu and Wei-Lin Chiang and
               Tianhao Wu and Joseph E. Gonzalez and M. Waleed Kadous and Ion Stoica},
  title     = {{RouteLLM}: Learning to Route {LLMs} with Preference Data},
  booktitle = {International Conference on Learning Representations ({ICLR})},
  publisher = {OpenReview.net},
  year      = {2025},
  url       = {https://openreview.net/forum?id=8sSqNntaMr}
}

@article{frugalgpt,
  author  = {Lingjiao Chen and Matei Zaharia and James Zou},
  title   = {{FrugalGPT}: How to Use Large Language Models While Reducing Cost and
             Improving Performance},
  journal = {Transactions on Machine Learning Research},
  volume  = {2024},
  year    = {2024},
  url     = {https://openreview.net/forum?id=cSimKw5p6R}
}

@inproceedings{masrouter,
  title     = {{MasRouter}: Learning to Route {LLMs} for Multi-Agent Systems},
  author    = {Yue, Yanwei and Zhang, Guibin and Liu, Boyang and Wan, Guancheng and
               Wang, Kun and Cheng, Dawei and Qi, Yiyan},
  booktitle = {Proceedings of the 63rd Annual Meeting of the Association for
               Computational Linguistics (Volume 1: Long Papers)},
  year      = {2025},
  publisher = {Association for Computational Linguistics},
  url       = {https://aclanthology.org/2025.acl-long.757/},
  doi       = {10.18653/v1/2025.acl-long.757},
  pages     = {15549--15572}
}

@inproceedings{routerdc,
  author    = {Shuhao Chen and Weisen Jiang and Baijiong Lin and James T. Kwok and
               Yu Zhang},
  title     = {{RouterDC}: Query-Based Router by Dual Contrastive Learning for
               Assembling Large Language Models},
  booktitle = {Advances in Neural Information Processing Systems 37 ({NeurIPS} 2024)},
  year      = {2024},
  url       = {http://papers.nips.cc/paper_files/paper/2024/hash/7a641b8ec86162fc875fb9f6456a542f-Abstract-Conference.html}
}

@inproceedings{graphrouter,
  author    = {Tao Feng and Yanzhen Shen and Jiaxuan You},
  title     = {{GraphRouter}: {A} Graph-based Router for {LLM} Selections},
  booktitle = {International Conference on Learning Representations ({ICLR})},
  publisher = {OpenReview.net},
  year      = {2025},
  url       = {https://openreview.net/forum?id=eU39PDsZtT}
}

@inproceedings{agentrouter,
    title     = {{AgentRouter}: A Knowledge-Graph-Guided {LLM} Router for
                 Collaborative Multi-Agent Question Answering},
    author    = {Zhang, Zheyuan and Shi, Kaiwen and Yuan, Zhengqing and Wang, Zehong
                 and Ma, Tianyi and Murugesan, Keerthiram and Galassi, Vincent and
                 Zhang, Chuxu and Ye, Yanfang},
    booktitle = {Proceedings of the 64th Annual Meeting of the Association for
                 Computational Linguistics (Volume 1: Long Papers)},
    year      = {2026},
    publisher = {Association for Computational Linguistics},
    url       = {https://aclanthology.org/2026.acl-long.33/},
    doi       = {10.18653/v1/2026.acl-long.33},
    pages     = {788--809}
}

@inproceedings{progrouter,
  title     = {{ProgRouter}: Online Progress-Guided Orchestration for Multi-Agent
               {LLM} Workflows under Quality-Cost Tradeoffs},
  author    = {Li, Songyuan and Abdelmoniem, Ahmed M. and Wang, Shiqiang},
  booktitle = {Findings of the Association for Computational Linguistics:
               {EMNLP} 2026},
  year      = {2026},
  publisher = {Association for Computational Linguistics},
  url       = {https://arxiv.org/abs/2608.25992}
}

@inproceedings{bamas,
  author    = {Yang, Liming and Luo, Junyu and Liu, Xuanzhe and Lou, Yiling and
               Chen, Zhenpeng},
  title     = {{BAMAS}: Structuring Budget-Aware Multi-Agent Systems},
  booktitle = {Proceedings of the AAAI Conference on Artificial Intelligence},
  volume    = {40},
  number    = {35},
  pages     = {29802--29810},
  year      = {2026},
  doi       = {10.1609/aaai.v40i35.40226}
}

@article{agentbalance,
  author        = {Shuowei Cai and Yansong Ning and Hao Liu},
  title         = {{AgentBalance}: Backbone-then-Topology Design for Cost-Effective
                   Multi-Agent Systems under Budget Constraints},
  journal       = {CoRR},
  volume        = {abs/2512.11426},
  year          = {2025},
  url           = {https://arxiv.org/abs/2512.11426},
  eprinttype    = {arXiv},
  eprint        = {2512.11426}
}

@inproceedings{cot,
  author    = {Jason Wei and Xuezhi Wang and Dale Schuurmans and Maarten Bosma and
               Brian Ichter and Fei Xia and Ed H. Chi and Quoc V. Le and Denny Zhou},
  title     = {Chain-of-Thought Prompting Elicits Reasoning in Large Language Models},
  booktitle = {Advances in Neural Information Processing Systems 35 ({NeurIPS} 2022)},
  year      = {2022},
  url       = {http://papers.nips.cc/paper_files/paper/2022/hash/9d5609613524ecf4f15af0f7b31abca4-Abstract-Conference.html}
}

@inproceedings{react,
  author    = {Shunyu Yao and Jeffrey Zhao and Dian Yu and Nan Du and Izhak Shafran and
               Karthik Narasimhan and Yuan Cao},
  title     = {{ReAct}: Synergizing Reasoning and Acting in Language Models},
  booktitle = {International Conference on Learning Representations ({ICLR})},
  publisher = {OpenReview.net},
  year      = {2023},
  url       = {https://openreview.net/forum?id=WE_vluYUL-X}
}

@inproceedings{reflexion,
  author    = {Noah Shinn and Federico Cassano and Edward Berman and Ashwin Gopinath and
               Karthik Narasimhan and Shunyu Yao},
  title     = {Reflexion: Language Agents with Verbal Reinforcement Learning},
  booktitle = {Advances in Neural Information Processing Systems 36 ({NeurIPS} 2023)},
  year      = {2023},
  url       = {https://proceedings.neurips.cc/paper_files/paper/2023/hash/1b44b878bb782e6954cd888628510e90-Abstract-Conference.html}
}

@inproceedings{chatdev,
  author    = {Chen Qian and Wei Liu and Hongzhang Liu and Nuo Chen and Yufan Dang and
               Jiahao Li and Cheng Yang and Weize Chen and Yusheng Su and Xin Cong and
               Juyuan Xu and Dahai Li and Zhiyuan Liu and Maosong Sun},
  title     = {{ChatDev}: Communicative Agents for Software Development},
  booktitle = {Proceedings of the 62nd Annual Meeting of the Association for
               Computational Linguistics (Volume 1: Long Papers)},
  publisher = {Association for Computational Linguistics},
  year      = {2024},
  url       = {https://aclanthology.org/2024.acl-long.810/},
  doi       = {10.18653/v1/2024.acl-long.810},
  pages     = {15174--15186}
}

@inproceedings{agentverse,
  author    = {Weize Chen and Yusheng Su and Jingwei Zuo and Cheng Yang and Chenfei Yuan and
               Chi{-}Min Chan and Heyang Yu and Yaxi Lu and Yi{-}Hsin Hung and Chen Qian and
               Yujia Qin and Xin Cong and Ruobing Xie and Zhiyuan Liu and Maosong Sun and
               Jie Zhou},
  title     = {AgentVerse: Facilitating Multi-Agent Collaboration and Exploring Emergent
               Behaviors},
  booktitle = {International Conference on Learning Representations ({ICLR})},
  publisher = {OpenReview.net},
  year      = {2024},
  url       = {https://openreview.net/forum?id=EHg5GDnyq1}
}

@inproceedings{moa,
  author    = {Junlin Wang and Jue Wang and Ben Athiwaratkun and Ce Zhang and James Zou},
  title     = {Mixture-of-Agents Enhances Large Language Model Capabilities},
  booktitle = {International Conference on Learning Representations ({ICLR})},
  publisher = {OpenReview.net},
  year      = {2025},
  url       = {https://openreview.net/forum?id=h0ZfDIrj7T}
}

@inproceedings{critic,
  author    = {Zhibin Gou and Zhihong Shao and Yeyun Gong and Yelong Shen and Yujiu Yang and
               Nan Duan and Weizhu Chen},
  title     = {{CRITIC}: Large Language Models Can Self-Correct with Tool-Interactive
               Critiquing},
  booktitle = {International Conference on Learning Representations ({ICLR})},
  publisher = {OpenReview.net},
  year      = {2024},
  url       = {https://openreview.net/forum?id=Sx038qxjek}
}

@inproceedings{planningabilities,
  author    = {Karthik Valmeekam and Matthew Marquez and Sarath Sreedharan and
               Subbarao Kambhampati},
  title     = {On the Planning Abilities of Large Language Models - {A} Critical
               Investigation},
  booktitle = {Advances in Neural Information Processing Systems 36 ({NeurIPS} 2023)},
  year      = {2023},
  url       = {https://proceedings.neurips.cc/paper_files/paper/2023/hash/efb2072a358cefb75886a315a6fcf880-Abstract-Conference.html}
}

@inproceedings{contextfolding,
  author    = {Weiwei Sun and Miao Lu and Zhan Ling and Kang Liu and Xuesong Yao and
               Yiming Yang and Jiecao Chen},
  title     = {Scaling Long-Horizon Agent via Context Folding},
  booktitle = {Proceedings of the 43rd International Conference on Machine Learning},
  publisher = {PMLR},
  year      = {2026},
  url       = {https://openreview.net/forum?id=lNRgWoGfYg}
}

@inproceedings{puppets,
  author    = {Yufan Dang and Chen Qian and Xueheng Luo and Jingru Fan and
               Zihao Xie and Ruijie Shi and Weize Chen and Cheng Yang and
               Xiaoyin Che and Ye Tian and Xuantang Xiong and Lei Han and
               Zhiyuan Liu and Maosong Sun},
  title     = {Multi-Agent Collaboration via Evolving Orchestration},
  booktitle = {Advances in Neural Information Processing Systems 38 ({NeurIPS} 2025)},
  year      = {2025},
  url       = {http://papers.nips.cc/paper_files/paper/2025/hash/f1320d2e2842169c6fc89dcbd80e94d0-Abstract-Conference.html}
}

@inproceedings{autogen,
  author    = {Qingyun Wu and Gagan Bansal and Jieyu Zhang and Yiran Wu and
               Beibin Li and Erkang Zhu and Li Jiang and Xiaoyun Zhang and
               Shaokun Zhang and Jiale Liu and Ahmed Hassan Awadallah and
               Ryen W. White and Doug Burger and Chi Wang},
  title     = {{AutoGen}: Enabling Next-Gen {LLM} Applications via Multi-Agent
               Conversation},
  booktitle = {First Conference on Language Modeling, {COLM} 2024},
  year      = {2024},
  url       = {https://openreview.net/forum?id=uAjxFFing2}
}

@inproceedings{metagpt,
  author    = {Sirui Hong and Mingchen Zhuge and Jonathan Chen and Xiawu Zheng and
               Yuheng Cheng and Jinlin Wang and Ceyao Zhang and Zili Wang and
               Steven Ka Shing Yau and Zijuan Lin and Liyang Zhou and Chenyu Ran and
               Lingfeng Xiao and Chenglin Wu and J{\"u}rgen Schmidhuber},
  title     = {{MetaGPT}: Meta Programming for {A} Multi-Agent Collaborative
               Framework},
  booktitle = {International Conference on Learning Representations ({ICLR})},
  publisher = {OpenReview.net},
  year      = {2024},
  url       = {https://openreview.net/forum?id=VtmBAGCN7o}
}

@inproceedings{llmdebate,
  author    = {Yilun Du and Shuang Li and Antonio Torralba and Joshua B. Tenenbaum and
               Igor Mordatch},
  title     = {Improving Factuality and Reasoning in Language Models through
               Multiagent Debate},
  booktitle = {Proceedings of the 41st International Conference on Machine
               Learning ({ICML})},
  series    = {Proceedings of Machine Learning Research},
  volume    = {235},
  pages     = {11733--11763},
  publisher = {PMLR},
  year      = {2024},
  url       = {https://proceedings.mlr.press/v235/du24e.html}
}

@inproceedings{macnet,
  author    = {Chen Qian and Zihao Xie and Yifei Wang and Wei Liu and Kunlun Zhu and
               Hanchen Xia and Yufan Dang and Zhuoyun Du and Weize Chen and Cheng Yang and
               Zhiyuan Liu and Maosong Sun},
  title     = {Scaling Large Language Model-based Multi-Agent Collaboration},
  booktitle = {International Conference on Learning Representations ({ICLR})},
  publisher = {OpenReview.net},
  year      = {2025},
  url       = {https://openreview.net/forum?id=K3n5jPkrU6}
}

@inproceedings{treacle,
  author    = {Xuechen Zhang and Zijian Huang and Ege Onur Taga and Carlee Joe-Wong and
               Samet Oymak and Jiasi Chen},
  title     = {Efficient Contextual {LLM} Cascades through Budget-Constrained Policy
               Learning},
  booktitle = {Advances in Neural Information Processing Systems 37 ({NeurIPS} 2024)},
  year      = {2024},
  url       = {https://proceedings.neurips.cc/paper_files/paper/2024/hash/a6deba3b2408af45b3f9994c2152b862-Abstract-Conference.html}
}

@inproceedings{generativeagents,
  author    = {Joon Sung Park and Joseph C. O'Brien and Carrie Jun Cai and
               Meredith Ringel Morris and Percy Liang and Michael S. Bernstein},
  title     = {Generative Agents: Interactive Simulacra of Human Behavior},
  booktitle = {Proceedings of the 36th Annual {ACM} Symposium on User Interface Software
               and Technology ({UIST})},
  publisher = {ACM},
  year      = {2023},
  url       = {https://doi.org/10.1145/3586183.3606763},
  doi       = {10.1145/3586183.3606763},
  pages     = {2:1--2:22}
}

@inproceedings{tokeneconomies,
  author    = {Junlin Wang and Siddhartha Jain and Dejiao Zhang and Baishakhi Ray and
               Varun Kumar and Ben Athiwaratkun},
  title     = {Reasoning in Token Economies: Budget-Aware Evaluation of {LLM} Reasoning
               Strategies},
  booktitle = {Proceedings of the Conference on Empirical Methods in Natural Language
               Processing ({EMNLP})},
  publisher = {Association for Computational Linguistics},
  year      = {2024},
  url       = {https://aclanthology.org/2024.emnlp-main.1112/},
  doi       = {10.18653/v1/2024.emnlp-main.1112},
  pages     = {19916--19939}
}

@inproceedings{toolformer,
  author    = {Timo Schick and Jane Dwivedi-Yu and Roberto Dess{\`\i} and Roberta Raileanu and
               Maria Lomeli and Eric Hambro and Luke Zettlemoyer and Nicola Cancedda and
               Thomas Scialom},
  title     = {Toolformer: Language Models Can Teach Themselves to Use Tools},
  booktitle = {Advances in Neural Information Processing Systems 36 ({NeurIPS} 2023)},
  year      = {2023},
  url       = {https://proceedings.neurips.cc/paper_files/paper/2023/hash/d842425e4bf79ba039352da0f658a906-Abstract-Conference.html}
}

@inproceedings{toolllm,
  author    = {Yujia Qin and Shihao Liang and Yining Ye and Kunlun Zhu and Lan Yan and
               Yaxi Lu and Yankai Lin and Xin Cong and Xiangru Tang and Bill Qian and
               Sihan Zhao and Lauren Hong and Runchu Tian and Ruobing Xie and Jie Zhou and
               Mark Gerstein and Dahai Li and Zhiyuan Liu and Maosong Sun},
  title     = {{ToolLLM}: Facilitating Large Language Models to Master 16000+ Real-world
               {APIs}},
  booktitle = {International Conference on Learning Representations ({ICLR})},
  publisher = {OpenReview.net},
  year      = {2024},
  url       = {https://openreview.net/forum?id=dHng2O0Jjr}
}

@article{c2mabv,
  author        = {Xiangxiang Dai and Jin Li and Xutong Liu and Anqi Yu and
                   John C. S. Lui},
  title         = {Cost-Effective Online Multi-{LLM} Selection with Versatile
                   Reward Models},
  journal       = {CoRR},
  volume        = {abs/2405.16587},
  year          = {2024},
  url           = {https://arxiv.org/abs/2405.16587},
  eprinttype    = {arXiv},
  eprint        = {2405.16587}
}

@misc{drgrpo,
      title={Understanding R1-Zero-Like Training: A Critical Perspective}, 
      author={Zichen Liu and Changyu Chen and Wenjun Li and Penghui Qi and Tianyu Pang and Chao Du and Wee Sun Lee and Min Lin},
      year={2025},
      eprint={2503.20783},
      archivePrefix={arXiv},
      primaryClass={cs.LG},
      url={https://arxiv.org/abs/2503.20783}, 
}

@misc{artificialanalysis_models,
  author       = {{Artificial Analysis}},
  title        = {Artificial Analysis: Models},
  year         = {2026},
  howpublished = {\url{https://artificialanalysis.ai/models/}},
  note         = {Accessed: 2026-07-17}
}

@article{omnimath2,
  author        = {Marthe Ballon and Andres Algaba and Brecht Verbeken and
                   Vincent Ginis},
  title         = {Benchmarks Saturate When The Model Gets Smarter Than The Judge},
  journal       = {CoRR},
  volume        = {abs/2601.19532},
  year          = {2026},
  url           = {https://arxiv.org/abs/2601.19532},
  eprinttype    = {arXiv},
  eprint        = {2601.19532}
}

@inproceedings{mmlupro,
  author    = {Yubo Wang and Xueguang Ma and Ge Zhang and Yuansheng Ni and
               Abhranil Chandra and Shiguang Guo and Weiming Ren and Aaran Arulraj
               and Xuan He and Ziyan Jiang and Tianle Li and Max Ku and Kai Wang
               and Alex Zhuang and Rongqi Fan and Xiang Yue and Wenhu Chen},
  title     = {{MMLU-Pro}: {A} More Robust and Challenging Multi-Task Language
               Understanding Benchmark},
  booktitle = {Advances in Neural Information Processing Systems 37 ({NeurIPS} 2024),
               Datasets and Benchmarks Track},
  year      = {2024},
  url       = {https://proceedings.neurips.cc/paper_files/paper/2024/hash/ad236edc564f3e3156e1b2feafb99a24-Abstract-Datasets_and_Benchmarks_Track.html}
}

@inproceedings{musr,
  author    = {Zayne Sprague and Xi Ye and Kaj Bostrom and Swarat Chaudhuri and
               Greg Durrett},
  title     = {{MuSR}: Testing the Limits of Chain-of-thought with Multistep Soft
               Reasoning},
  booktitle = {International Conference on Learning Representations ({ICLR})},
  publisher = {OpenReview.net},
  year      = {2024},
  url       = {https://openreview.net/forum?id=jenyYQzue1}
}

@inproceedings{bbeh,
  author    = {Mehran Kazemi and Bahare Fatemi and Hritik Bansal and John Palowitch
               and Chrysovalantis Anastasiou and Sanket Vaibhav Mehta and
               Lalit K. Jain and Virginia Aglietti and Disha Jindal and Peter Chen
               and Nishanth Dikkala and Gladys Tyen and Xin Liu and Uri Shalit and
               Silvia Chiappa and Kate Olszewska and Yi Tay and Vinh Q. Tran and
               Quoc V. Le and Orhan Firat},
  title     = {{BIG-Bench} Extra Hard},
  booktitle = {Proceedings of the 63rd Annual Meeting of the Association for
               Computational Linguistics (Volume 1: Long Papers)},
  publisher = {Association for Computational Linguistics},
  year      = {2025},
  url       = {https://aclanthology.org/2025.acl-long.1285/},
  doi       = {10.18653/v1/2025.acl-long.1285},
  pages     = {26473--26501}
}

@inproceedings{evalplus,
  author    = {Jiawei Liu and Chunqiu Steven Xia and Yuyao Wang and Lingming Zhang},
  title     = {Is Your Code Generated by {ChatGPT} Really Correct? Rigorous
               Evaluation of Large Language Models for Code Generation},
  booktitle = {Advances in Neural Information Processing Systems 36 ({NeurIPS} 2023)},
  year      = {2023},
  url       = {https://proceedings.neurips.cc/paper_files/paper/2023/hash/43e9d647ccd3e4b7b5baab53f0368686-Abstract-Conference.html}
}

@article{wang2020minilm,
  title={Minilm: Deep self-attention distillation for task-agnostic compression of pre-trained transformers},
  author={Wang, Wenhui and Wei, Furu and Dong, Li and Bao, Hangbo and Yang, Nan and Zhou, Ming},
  journal={Advances in neural information processing systems},
  volume={33},
  pages={5776--5788},
  year={2020}
}
\endgroup

\clearpage
\appendix

\section{Notations}
\label{app:notation}
We conclude the used notations in table~\ref{tab:notation}

{
\begin{longtable}{lp{0.68\columnwidth}}
\caption{Summary of notation.}\label{tab:notation}\\
\toprule
\textbf{Symbol} & \textbf{Description} \\
\midrule
\endfirsthead
\toprule
\textbf{Symbol} & \textbf{Description} \\
\midrule
\endhead
\midrule
\multicolumn{2}{r}{\textit{Continued on next page}} \\
\endfoot
\bottomrule
\endlastfoot
\multicolumn{2}{l}{\textit{Input and trajectory}} \\
$\mathcal{Q}$ & User query. \\ 
$B$ & Cost budget. \\
$\mathcal{P}$ & Distribution of query--budget pairs. \\ 
$\xi$ & Multi-agent execution trajectory. \\ 
$N$ & Number of agents. \\
$[B_{\min},B_{\max}]$ & Predefined training budget range. \\
$y_i$ & Output message of agent $i$. \\
$\tilde y_i$ & Message agent $i$ transmits to its successors. \\
$T^{\mathrm{in}}_i,\ T^{\mathrm{out}}_i$ & Input and output token counts of agent $i$. \\
$\bytes{\cdot},\ |\cdot|_{\mathrm{chars}}$ & Size in bytes; length in characters. \\
$o_i$ & Observation of agent $i$, $\big(\bytes{\tilde y_i},T^{\mathrm{in}}_i,|y_i|_{\mathrm{chars}},T^{\mathrm{out}}_i\big)$. \\
$o_{<i}$ & Observations of agents $1,\dots,i-1$. \\

\midrule
\multicolumn{2}{l}{\textit{Search space and graph}} \\ 
$\mathcal{R},\ N_r$ & Pool of candidate roles and its size. \\
$\mathcal{M},\ N_m$ & Pool of backbone LLMs and its size. \\
$r_i$ & Agent role. \\
$m_i$ & Backbone language model. \\
$c^{\mathrm{in}}_m,\ c^{\mathrm{out}}_m$ & Price (in USD\$) per input and output token of backbone $m$. \\
$\underline{m},\ \bar m$ & Cheapest backbone; mid-priced backbone. \\
$\mathcal{G}=(\mathcal{V},\mathcal{E})$ & Multi-agent system as a DAG. \\
$v_i=\{r_i,m_i\}$ & Agent $i$. \\
$e_{ji}$ & Edge from agent $j$ to agent $i$. \\
$\mathbf{e}_{>j,i}$ & Edge decisions of agent $i$ already made in the reverse scan. \\
$\mathcal{S}$ & Space of agent states. \\
$\mathcal{N}_{in}(v_i)$ & Direct predecessors of agent $i$. \\
$\mathcal{G}_{<i}$ & Partial graph of the first $i-1$ agents. \\
$N_{\max}$ & Maximum number of agents. \\
$\mathcal{T},\ t$ & Pool of topology templates; a template. \\

\midrule
\multicolumn{2}{l}{\textit{Policy decisions}} \\
$b_i$ & Remaining budget. \\
$s_i=\left(
\mathcal{Q},B,
b_i,
\mathcal{G}_{<i},
o_{<i}
\right)$ & Policy state before step $i$. \\
$\tau_i$ & Termination decision. \\
$\pi_\theta$ & Routing policy parameterized by $\theta$. \\
$P_\theta(\mathcal{G}\mid\cdot)$ & Policy-induced distribution over graphs. \\
$\mathcal{I}$ & Construction steps actually reached. \\
$I^\tau_i,\ I^e_{ji},\ I^r_i$ & Indicators of termination, edge and role decisions sampled from $\pi_\theta$. \\
$\mathcal{M}^m_i$ & Affordable backbone set at step $i$. \\

\midrule
\multicolumn{2}{l}{\textit{Optimization}} \\
$S(\mathcal{Q},\xi)$ & Task performance (reward). \\
$C_i$ & Cost of agent $i$. \\
$C(\xi)$ & Total trajectory cost. \\ 
$U(\mathcal{Q},B,\xi)$ & Utility, $S(\mathcal{Q},\xi)-\gamma\,C(\xi)/B$. \\
$\gamma$ & Cost-sensitivity coefficient. \\

\midrule
\multicolumn{2}{l}{\textit{Model architecture}} \\ 
$\Enc(\cdot),\ \LN(\cdot)$ & Frozen sentence encoder; layer normalization. \\
$D,\ d$ & Encoder output dimension; hidden dimension. \\
$f_\mathcal{Q},\ f_B$ & Query embedding; log-budget embedding. \\
$\tilde f_\mathcal{Q}$ & Budget-conditioned query embedding. \\
$z_r,\ \mathbf{Z}$ & Role embedding; role embedding matrix. \\
$z_m$ & Backbone embedding. \\
$\mathbf{l}_m$ & Standardized log prices $(\ell^{\mathrm{in}}_m,\ell^{\mathrm{out}}_m)$ of backbone $m$. \\
$f^{(i)}_{\mathrm{hist}}$ & History embedding of generated agents. \\
$\phi_i$ & Cost vector. \\
$\psi(\cdot),\ \boldsymbol{\psi}(B,b_i)$ & Rescaled log-dollar map; the pair $(\psi(B),\psi(b_i))$. \\
$f^{(i)}_{\mathrm{cont}},\ g_i$ & Context embedding; its gate. \\
$f^{(i)}_{\mathrm{edge}}$ & Connectivity embedding of previous agents. \\
$h^{(i)}_{\mathrm{node}}$ & Node-level hidden state. \\
$h^{(i,j)}_{\mathrm{edge}}$ & Edge-level hidden state when scanning agent $j$ in agent selecting state $i$. \\
$\boldsymbol{\ell}_{j\to i}$ & Length vector of edge $(v_j,v_i)$. \\
$\omega^{i}_{\mathrm{node}},\ \omega^{i}_{\mathrm{m}},\ \omega^{i,j}_{\mathrm{edge}}$ & Role, backbone and edge score. \\
$\mathrm{GRU}_{(\cdot)}$ & Gated Recurrent Unit. \\
$\MLP_{(\cdot)}$ & Multilayer Perceptron. \\
$\sigma(\cdot)$ & Sigmoid function. \\
$\mathbb{I}[\cdot]$ & Indicator function \\

\midrule
\multicolumn{2}{l}{\textit{Meter}} \\ %
$\widehat{T}_i$ & Upper estimation of the input tokens of agent $i$. \\
$q_{\min}$ & Minimum admissible completion length, see table~\ref{tab:estimator}. \\
$F_m(T)$ & Floor cost of a call on backbone $m$. \\
$R_i$ & Reserved floor cost for successors. \\
$q_i$ & Completion cap of agent $i$. \\
$T_{\max}$ & Maximum completion tokens per call, see table~\ref{tab:estimator}. \\
$\mu_i$ & Marginal price per output token of agent $i$. \\
$\nu_i$ & Whether a successor is still obliged. \\
$\kappa_{\mathrm{r}}$ & Output-to-downstream-input token factor, see table~\ref{tab:estimator}. \\
$\hat\rho_i$ & Estimated input charge. \\

\midrule
\multicolumn{2}{l}{\textit{Stage 1: imitation}} \\
$\mathcal{D}_{\mathrm{SFT}}$ & Demonstration set (train and val splits). \\
$\mathfrak{B}$ & Log-spaced budget grid. \\
$\mathcal{R}_1(\mathcal{Q})$ & Roles the policy can produce for the first agent. \\
$K$ & Runs per template and query. \\
$\hat u_t,\ \hat s_t(q), \ \hat c_t$ & Empirical utility, mean success of template $t$, and mean cost of template $t$. \\
$t^\star$ & Selected template. \\
$\tau^\star_i,\ r^\star_i,\ e^\star_{ji}$ & Logged target decisions. \\
$\lambda_{\mathrm{stop}},\lambda_{\mathrm{role}},\lambda_{\mathrm{edge}}$ & Loss weights. \\
$w_r$ & Final-role weight of role $r$. \\
$\mathcal{S}_r,\ \mathcal{Q}_r$ & Runs with final role $r$; their queries. \\
$\mathcal{L}_{\mathrm{SFT}},\ \mathcal{L}_{\mathrm{val}}$ & Imitation loss; validation loss. \\
$\Delta_r,\ \tilde\Delta_r$ & Final-role effect; its shrunk version. \\
$\zeta,\ [w_-,w_+]$ & Temperature and clipping range of $w_r$. \\
$[\mathrm{lo}_r,\mathrm{hi}_r]$ & 95\% bootstrap interval of $\Delta_r$. \\
$E,\ P$ & Epochs; patience. \\
$N_e$ & Number of agents in demonstration $e$. \\
$Z$ & Mean number of decisions per demonstration. \\

\midrule
\multicolumn{2}{l}{\textit{Stage 2: reinforcement learning}} \\
$G$ & Group size. \\
$\theta_{\mathrm{old}}$ & Parameters of the sampling policy. \\
$U_g,\ A_g$ & Utility and advantage of trajectory $\xi_g$. \\
$\mathcal{K}(\xi),\ d$ & Policy decisions of $\xi$; a decision. \\
$\rho_d(\theta)$ & Likelihood ratio of decision $d$. \\
$\epsilon$ & Clipping range. \\
$\mathcal{L}_{\mathrm{RL}}$ & Clipped surrogate loss. \\
$N_{\mathrm{rd}},\ E,\ \beta$ & Rounds; epochs per round; minibatch. \\
\end{longtable}
}

\section{Supplementary Functions}
\label{app:realize}
\paragraph{History embedding.}
\begin{equation}
f^{(i)}_{\mathrm{hist}}=\mathrm{GRU}\Big(\big[[z_{r_1},z_{m_1}],\dots,[z_{r_{i-1}},z_{m_{i-1}}]\big]\Big), \,\,\, f^{(1)}_{\mathrm{hist}}=\mathbf{0}.
\label{eq:hist}
\end{equation}
The GRU reads the role and backbone embeddings of the agents recruited so far, in generation order, and summarizes the team built up to step $i$. It is zero before the first agent is generated.

\paragraph{Connectivity of the previous agent.}
\begin{equation}
f^{(i)}_{\mathrm{edge}}=\MLP_{\mathrm{struct}}\big([e_{1,i-1},\dots,e_{N_{\max},i-1},\;\mathbb{I}[1<i-1],\dots,\mathbb{I}[N_{\max}<i-1]]\big), \,\,\, \forall j\ge i-1 \,\, e_{j,i-1}:=0,
\label{eq:edge_emb}
\end{equation}
The first $N_{\max}$ entries are the incoming edges $\mathbf{e}_{\cdot i-1}$ of the previous agent $v_{i-1}$, and the last $N_{\max}$ entries mark which agents it scanned as candidate predecessors, namely $v_1,\dots,v_{i-2}$. Padding with zeros gives the input a fixed length $2N_{\max}$ at every step. It is the zero vector for $i\le2$, since $v_1$ has no predecessors.

\paragraph{Budget position.}
\begin{equation}
\psi(x)=\clip\!\left(
  \frac{2\big(\log\max\{x,\,10^{-3}B_{\min}\}-\log B_{\min}\big)}
       {\log B_{\max}-\log B_{\min}}-1,\;-3,\;1\right),
\qquad x\in\{B,\,b_i\}.
\label{eq:psi_b}
\end{equation}
The same function places the assigned budget $B$ and the remainder $b_i$ on one log-dollar scale. The training range $[B_{\min},B_{\max}]$ maps linearly in $\log v$ onto $[-1,1]$. Below $B_{\min}$ the scale extends down to $-3$, which lets a nearly exhausted remainder stay distinguishable. The floor $10^{-3}B_{\min}$ keeps the logarithm finite when $b_i$ reaches zero or turns negative after an overrun.

\paragraph{Length vector.}
\begin{equation}
\boldsymbol{\ell}_{j\to i}=\big(\log(1+|y_j|_{\mathrm{chars}}),\;\log(1+T^{\mathrm{out}}_j),\;\log(1+T^{\mathrm{in}}_j),\;\log c^{\mathrm{in}}_{m_i}\big).
\label{eq:length}
\end{equation}
When the edge head decides whether $v_i$ should read $v_j$, it sees how long $v_j$'s response turned out to be, how many tokens $v_j$ consumed and produced, and the input price of the receiving backbone $m_i$. Together these describe what connecting $v_j$ would cost $v_i$ to read. The first three entries come from the observation $o_j$ and are known only after $v_j$ has been executed. Logarithms compress lengths that vary over orders of magnitude.

\paragraph{Upper estimate of Input.}
$\widehat{T}_i=\lceil\bytes{q}/\kappa_{\mathrm{b}}\rceil+\kappa_{\mathrm{p}}
+\sum_{j<i}\big(\lceil\bytes{\tilde y_j}/\kappa_{\mathrm{b}}\rceil+\kappa_{\mathrm{m}}\big)$ be an
upper estimate of the input node $i$ would consume if it read \emph{every} completed predecessor.

\paragraph{Reserve.}
$R_i=\nu_i\,F_{\underline m}\big(\widehat{T}^{\mathrm{res}}_i\big)$ where
$\nu_i=\Ind[i]$ the system would not satisfy the set of nodes with no outgoing edge after node $i$ and $\widehat{T}^{\mathrm{res}}_i$ is the input the obliged successor would consume. Under $N_{\min}=1$ the reserve vanishes in the mask of Eq. \ref{eq:mask}: with at least one agent already executed no successor is structurally required before the mask is consulted. It binds in the completion cap of Eq. \ref{eq:cap}, where a node that leaves two sinks obliges a merge.

\paragraph{Estimated Input Charge.}
$\hat\rho_i=c^{\mathrm{in}}_{m_i}\big(\lceil\bytes{x_i}/\kappa_{\mathrm{b}}\rceil
+\kappa_{\mathrm{c}}\big)$ the estimated input charge.

\begin{table}[h]
\centering
\caption{Estimator constants and training hyperparameters. $\kappa_{\mathrm{r}}=\beta_{\mathrm{out}}/\kappa_{\mathrm{b}}$ with
$\beta_{\mathrm{out}}=8$ reserved output bytes per token. Choosing
$\kappa_{\mathrm{b}}=2$ rather than the customary $4$ makes the input estimate roughly twice the
expected token count; the guards are intended to be conservative, and the cost of that choice is a
slightly smaller usable completion cap. The imitation weights $\lambda$ are fixed in the SFT stage rather than read from the configuration.}
\begin{tabular}{lp{5.5cm}r l}
\toprule
Symbol & Role & Value & Config key \\
\midrule
\multicolumn{4}{@{}l}{\textit{Estimator}} \\
$\kappa_{\mathrm{b}}$ & UTF-8 bytes per estimated input token & $2$ & \texttt{input\_bytes\_per\_token} \\
$\kappa_{\mathrm{p}}$ & prompt reserve added to every mask estimate (tokens) & $1664$ & \texttt{prompt\_reserve\_tokens} \\
$\kappa_{\mathrm{c}}$ & chat-format overhead added to every request estimate & $768$ & \texttt{chat\_overhead\_tokens} \\
$\kappa_{\mathrm{m}}$ & per-predecessor margin in the mask estimate (tokens) & $256$ & --- \\
$\kappa_{\mathrm{r}}$ & input tokens a downstream reader pays per output token & $4$ & derived \\
$q_{\min}$ & minimum admissible completion & $256$ & \texttt{min\_completion\_tokens} \\
$T_{\max}$ & provider-side completion cap & $32768$ & \texttt{max\_tokens} \\
\midrule
\multicolumn{4}{@{}l}{\textit{Objective}} \\
$\gamma$ & weight of the budget share in the utility & $0.05$ & \texttt{budget\_share\_weight} \\
$\bar\varepsilon$ & cap on the overrun ratio & $1$ & \texttt{excess\_cap} \\
$B_{\min}$ & lower end of the budget range (\$) & $10^{-3}$ & \texttt{budget\_min} \\
$B_{\max}$ & upper end of the budget range (\$) & $5\times10^{-2}$ & \texttt{budget\_max} \\
\midrule
\multicolumn{4}{@{}l}{\textit{Training}} \\
$\lambda_{\stopk}$ & weight of the termination term in $\mathcal{L}_{\mathrm{SFT}}$ & $0.2$ & --- \\
$\lambda_{\rolek}$ & weight of the role term in $\mathcal{L}_{\mathrm{SFT}}$ & $0.2$ & --- \\
$\lambda_{\edgek}$ & weight of the edge term in $\mathcal{L}_{\mathrm{SFT}}$ & $0.8$ & --- \\
$G$ & trajectories per query--budget group & $8$ & \texttt{group\_size} \\
$\epsilon$ & clipping radius of the ratio in $\mathcal{L}_{\mathrm{RL}}$ & $0.2$ & \texttt{clip\_epsilon} \\
\bottomrule
\end{tabular}
\label{tab:estimator}
\end{table}

\section{Algorithm Workflow of MeterRouter}
\label{app:alg}
\begin{algorithm}[H]
\caption{Construction of $\mathcal{D}_{\mathrm{SFT}}$}
\label{alg:dsft}
\begin{algorithmic}[1]
\Require queries $\mathcal{D}$, templates $\mathcal{T}$, runs $K$, budget grid $\mathfrak{B}$, mid-priced backbone $\bar m$
\State $\mathcal{D}_{\mathrm{SFT}}\gets\emptyset$
\For{$\mathcal{Q}\in\mathcal{D}$}
  \For{$t\in\mathcal{T}$, $k=1,\dots,K$}
    \State $r_i\sim\mathrm{Unif}(\mathcal{R})$,\ $m_i\gets\bar m$ \ for every agent $v_i$ of $t$
    \State $\xi_{t,k}\gets\textsc{Execute}(t,\{r_i,m_i\};\mathcal{Q},B_{\max})$
  \EndFor
  \State $\hat c_t\gets\frac1K\sum_k C(\xi_{t,k})$,\ \ $t\in\mathcal{T}$
  \For{$B\in\mathfrak{B}$}
    \State $t^\star\gets\argmax_{t:\,\hat c_t\le B}\ \hat u_t(\mathcal{Q},B)=\frac{1}{K}\sum_{k=1}^{K}S(\mathcal{Q},\xi_{t,k})-\frac{\gamma}{B}\cdot\frac{1}{K}\sum_{k=1}^{K}C(\xi_{t,k})$ 
    \For{$k$ with $S(\mathcal{Q},\xi_{t^\star,k})=1$}
      \State $e\gets\big(\mathcal{Q},B,\mathcal{G}(\xi_{t^\star,k}),o_{1:N}\big)$,\ \ $b_i\gets B-\sum_{j<i}C_j$ \Comment{no re-execution}
      \State \textbf{if} $m_i\in\mathcal{M}^m_i\ \forall i$ \textbf{then} $\mathcal{D}_{\mathrm{SFT}}\gets\mathcal{D}_{\mathrm{SFT}}\cup\{e\}$
    \EndFor
  \EndFor
\EndFor
\State \Return $\mathcal{D}_{\mathrm{SFT}}$
\end{algorithmic}
\end{algorithm}

\begin{algorithm}[H]
\caption{Stage 1: budget-conditioned imitation}
\label{alg:stage1}
\begin{algorithmic}[1]
\Require $\mathcal{D}^{\mathrm{train}}_{\mathrm{SFT}}$, $\mathcal{D}^{\mathrm{val}}_{\mathrm{SFT}}$ (Algorithm~\ref{alg:dsft}), role weights $\{w_r\}$, epochs $E_{\max}$, patience $P$
\State freeze $\MLP^{\mathrm{GELU}}_{\mathrm{route}}$ and $\MLP_B$
\For{$\mathrm{epoch}=1,\dots,E_{\max}$}
  \For{minibatch $\beta\subset\{e_q\sim\mathrm{Unif}(\mathcal{D}^{\mathrm{train}}_{\mathrm{SFT}}(q))\}_q$}
    \State $\theta\gets\mathrm{Adam}\big(\theta,\nabla_\theta\hat{\mathcal{L}}_{\mathrm{SFT}}(\theta;\beta)\big)$ \Comment{Eq. \ref{eq:sft}}
  \EndFor
  \State \textbf{if} $\mathcal{L}_{\mathrm{val}}(\theta)$ is the best so far \textbf{then} $\theta_{\mathrm{SFT}}\gets\theta$
  \State \textbf{if} no improvement for $P$ epochs \textbf{then break}
\EndFor
\State $Z\gets\operatorname{mean}_e\big(3N_e-1+N_e(N_e-1)/2+\ind[N_e<N_{\max}]\big)$ 
\State \Return $\theta_{\mathrm{SFT}}$, $Z$
\end{algorithmic}
\end{algorithm}

Here, let $\mathcal{D}^{\mathrm{train}}_{\mathrm{SFT}}(q)$ be the demonstrations of query $q$ and $\mathcal{Q}^{\mathrm{SFT}}=\{q:\mathcal{D}^{\mathrm{train}}_{\mathrm{SFT}}(q)\neq\emptyset\}$.The imitation objective is:
\begin{equation}
\begin{aligned}
\mathcal{L}_{\mathrm{SFT}}(\theta)=-\,\mathbb{E}_{q\sim\Unif(\mathcal{Q}^{\mathrm{SFT}})}\,\mathbb{E}_{e\sim\Unif(\mathcal{D}^{\mathrm{train}}_{\mathrm{SFT}}(q))}\Bigg[\;
&\lambda_{\stopk}\sum_{i\in\mathcal{I}}I^\tau_i\log\pi_\theta(\tau^\star_i\mid s_i)\\
&+\lambda_{\rolek}\sum_{i=1}^{N}I^r_i\,w_{r_N}^{\ind[i=N]}\log\pi_\theta(r^\star_i\mid s_i)\\
&+\lambda_{\edgek}\sum_{i=1}^{N}\sum_{j<i}I^e_{ji}\log\pi_\theta\big(e^\star_{ji}\mid r_i,m_i,\mathbf{e}_{>j,i},s_i\big)\Bigg],
\end{aligned}
\label{eq:sft}
\end{equation}
where $I^r_i=\ind[i\ge2\ \vee\ r_1\in\mathcal{R}_1(\mathcal{Q})]$ drops a logged first role that $\pi_\theta$ cannot produce, and $w_r$ is the final-role weight of Eq. \ref{eq:roleweights}. The backbone factor of Eq. \ref{eq:factorization} is absent because $\MLP^{\mathrm{GELU}}_{\mathrm{route}}$ is frozen. The two expectations are the sampling scheme of algorithm~\ref{alg:stage1}: every query receives equal weight, however many budgets and runs it contributes.

Early stopping monitors the structural imitation loss on the validation demonstrations:

\begin{equation}
\begin{aligned}
\mathcal{L}_{\mathrm{val}}(\theta)=-\,\frac{1}{|\mathcal{D}^{\mathrm{val}}_{\mathrm{SFT}}|}\sum_{e\in\mathcal{D}^{\mathrm{val}}_{\mathrm{SFT}}}\Bigg[\;
&\lambda_{\stopk}\sum_{i\in\mathcal{I}}I^\tau_i\log\pi_\theta(\tau^\star_i\mid s_i)\\
&+\lambda_{\edgek}\sum_{i=1}^{N}\sum_{j<i}I^e_{ji}\log\pi_\theta\big(e^\star_{ji}\mid r_i,m_i,\mathbf{e}_{>j,i},s_i\big)\Bigg].
\end{aligned}
\label{eq:val}
\end{equation}

\noindent\textbf{Final-role weights.}
Within a query and a template, the $K$ runs differ only in their randomly drawn roles, so the collected runs contain a paired contrast between roles. For every training run $\xi_{q,t,k}$, let $\hat s_t(q)=\frac1K\sum_{k'}S(\mathcal{Q}_q,\xi_{q,t,k'})$, $u_{q,t,k}=2\big(S(\mathcal{Q}_q,\xi_{q,t,k})-\hat s_t(q)\big)$, and let $r_N(\xi)$ be the role of the final agent of $\xi$. For each role $r$, let $\mathcal{S}_r=\{(q,t,k):r_N(\xi_{q,t,k})=r\}$ and let $\mathcal{Q}_r$ be the queries appearing in $\mathcal{S}_r$. Then:
\begin{equation}
\begin{gathered}
\Delta_r=\frac{1}{|\mathcal{S}_r|}\sum_{(q,t,k)\in\mathcal{S}_r}u_{q,t,k},\qquad
\tilde\Delta_r=\operatorname{sgn}(\Delta_r)\Big(|\Delta_r|-\frac{\mathrm{hi}_r-\mathrm{lo}_r}{2}\Big)_+,\\
w_r=\frac{\clip\big(e^{\tilde\Delta_r/\zeta},w_-,w_+\big)}{\operatorname{mean}_{e\in\mathcal{D}^{\mathrm{train}}_{\mathrm{SFT}}}\clip\big(e^{\tilde\Delta_{r_N(e)}/\zeta},w_-,w_+\big)},
\end{gathered}
\label{eq:roleweights}
\end{equation}
where $[\mathrm{lo}_r,\mathrm{hi}_r]$ is the $95\%$ interval of $\Delta_r$ from a bootstrap that resamples the queries in $\mathcal{Q}_r$ with replacement, and $\tilde\Delta_r=0$ when $|\mathcal{Q}_r|<2$. With $K=2$, $u_{q,t,k}=S_k-S_{k'}\in\{-1,0,1\}$ compares the two runs of the same query and topology. Shrinking by the half-width of the interval keeps the unnormalized weight at one unless the effect of the final role exceeds its sampling uncertainty, and the normalization makes the average weight over the demonstrations equal to one. We use $\zeta=0.1$ and $[w_-,w_+]=[0.25,4]$.

\begin{algorithm}[H]
\caption{Stage 2: group-relative reinforcement learning}
\label{alg:stage2}
\begin{algorithmic}[1]
\Require $\theta_{\mathrm{SFT}}$, $Z$ (Algorithm~\ref{alg:stage1}), queries $\mathcal{D}$, group size $G$, rounds $N_{\mathrm{rd}}$, epochs $E$
\State $\theta\gets\theta_{\mathrm{SFT}}$
\For{$n=1,\dots,N_{\mathrm{rd}}$}
  \State $\theta_{\mathrm{old}}\gets\theta$
  \For{$\mathcal{Q}\in\mathcal{D}_n$} \Comment{queries of round $n$}
    \State $B\sim\mathrm{LogUnif}[B_{\min},B_{\max}]$
    \State $\xi_g\sim\pi_{\theta_{\mathrm{old}}}(\cdot\mid\mathcal{Q},B)$,\ \ $g=1,\dots,G$ \Comment{Algorithm~\ref{alg:inference}}
    \State $A_g\gets U(\mathcal{Q},B,\xi_g)-\frac1G\sum_{g'}U(\mathcal{Q},B,\xi_{g'})$ 
  \EndFor
  \For{$\mathrm{epoch}=1,\dots,E$;\ minibatch $\beta$ of trajectories}
    \State $\rho_d(\theta)\gets\pi_\theta(d)/\pi_{\theta_{\mathrm{old}}}(d)$,\ \ $d\in\mathcal{K}(\xi)$,\ $\xi\in\beta$
    \State $\theta\gets\mathrm{Adam}\big(\theta,\nabla_\theta\hat{\mathcal{L}}_{\mathrm{RL}}(\theta;\beta)\big)$ \Comment{Eq. \ref{eq:rl}}
  \EndFor
\EndFor
\State \Return $\theta$
\end{algorithmic}
\end{algorithm}

\begin{algorithm}[H]
\caption{Budget-aware inference}
\label{alg:inference}
\begin{algorithmic}[1]
\Require query $\mathcal{Q}$, budget $B$, trained policy $\pi_\theta$
\State $b_1\gets B$,\quad $\mathcal{G}\gets\emptyset$
\For{$i=1,\dots,N_{\max}+1$}
  \If{$i\ge2$ \textbf{and} $\mathcal{G}_{<i}$ is connected}
    \State \textbf{if} $i=N_{\max}+1$ \textbf{or} $\mathcal{M}^m_i=\emptyset$ \textbf{then break}
    \State $\tau_i\sim\pi_\theta(\tau_i\mid s_i)$ 
    \State \textbf{if} $\tau_i=1$ \textbf{then break}
  \EndIf
  \State $r_i\sim\pi_\theta(r_i\mid s_i)$ 
  \State $m_i\sim\pi_\theta(m_i\mid r_i,s_i)$ over $\mathcal{M}^m_i$ 
  \For{$j=i-1,\dots,1$}
    \State $e_{ji}\sim\pi_\theta(e_{ji}\mid r_i,m_i,\mathbf{e}_{>j,i},s_i)$ 
  \EndFor
  \State run $v_i$ on $m_i$, reading $\{\tilde y_j:e_{ji}=1\}$
  \State observe $o_i$;\quad $b_{i+1}\gets b_i-C_i$;\quad $s_{i+1}\gets(\mathcal{Q},B,b_{i+1},\mathcal{G}_{<i+1},o_{\le i})$ \label{alg:inf:meter}
\EndFor
\State \Return the response $y_N$ of the final agent $v_N$
\end{algorithmic}
\end{algorithm}
\section{Supplementary Details of Experiments}
\label{app:exp}

We conclude the dataset statistics in Table~\ref{tab:dataset-split}

\begin{table}[H]
\centering
\begin{minipage}{0.75\linewidth}
\caption{Dataset split used for training, validation, and testing.}
\label{tab:dataset-split}
\centering

\begin{tabular*}{\linewidth}{
    @{\extracolsep{\fill}} lccc @{}
}
\toprule
\textbf{Dataset}
& \textbf{Training}
& \textbf{Validation}
& \textbf{Test} \\
\midrule
OmniMath-2 & 200 & 40 & 1,000 \\
MMLU-Pro   & 200 & 40 & 1,000 \\
MuSR       & 200 & 40 & 516 \\
BBEH       & 200 & 40 & 1,000 \\
MBPP+      & 200 & 40 & 138 \\
\midrule
\textbf{Total}
& \textbf{1,000}
& \textbf{200}
& \textbf{3,654} \\
\bottomrule
\end{tabular*}
\end{minipage}
\end{table}

\begin{table}[H]
\centering
\begin{minipage}{0.90\linewidth}
\caption{LLM pool with each model's price tier, parameter counts, and
context length. Prices are the accounting rates used to compute costs,
taken from Artificial Analysis~\cite{artificialanalysis_models}.}
\label{tab:model-cards}
\centering

\begin{tabular*}{\linewidth}{
    @{\extracolsep{\fill}} lllrr @{}
}
\toprule
\textbf{LLM}
& \textbf{Tier}
& \textbf{Context}
& \multicolumn{2}{c}{\textbf{Price (\$/1M tokens)}} \\
\cmidrule(lr){4-5}
& & & \textbf{Input} & \textbf{Output} \\
\midrule
gpt-oss-20b
& Low-cost
& 128K
& 0.06
& 0.19 \\
gpt-oss-120b
& Mid-cost
& 128K
& 0.15
& 0.595 \\
MiniMax-M3 ($\sim$428B)
& High-cost
& 1M
& 0.30
& 1.20 \\
\bottomrule
\end{tabular*}
\end{minipage}
\end{table}

\subsection{Seed Variance of MeterRouter and MasRouter}
\label{app:seed}
MasRouter has the highest average accuracy among the baselines among MAS methods, so we trained and evaluated \texttt{MeterRouter} and MasRouter with three independent seeds on each benchmark in Table~\ref{tab:seed}

\begin{table}[H]
\centering
\caption{Accuracy (\%) and cost per query ($\times10^{-3}$ USD),
reported as mean $\pm$ standard deviation over three seeds.
\texttt{MeterRouter} uses a budget of $B=5\times10^{-2}$ USD.}
\label{tab:seed}
\resizebox{\linewidth}{!}{%
\begin{tabular}{lccccc}
\toprule
Method / Metric & \textbf{Omni-MATH-2} & \textbf{MMLU-Pro} & \textbf{MuSR} & \textbf{BBEH} & \textbf{MBPP+} \\
\midrule
MasRouter
& $67.5 \pm 1.3$
& $80.3 \pm 2.4$
& $67.9 \pm 2.2$
& $48.9 \pm 1.9$
& $80.8 \pm 2.7$ \\
\quad Cost
& $16.75 \pm 2.83$
& $3.92 \pm 0.23$
& $3.85 \pm 0.36$
& $25.63 \pm 3.02$
& $2.65 \pm 0.27$ \\
\midrule
\texttt{MeterRouter}
& $64.7 \pm 1.4$
& $83.4 \pm 2.5$
& $71.6 \pm 1.9$
& $45.4 \pm 1.8$
& $83.8 \pm 2.9$ \\
\quad Cost
& $4.12 \pm 0.45$
& $1.68 \pm 0.21$
& $0.61 \pm 0.08$
& $2.95 \pm 0.35$
& $0.30 \pm 0.02$ \\
\bottomrule
\end{tabular}%
}
\end{table}

\subsection{Single Backbone Ablation}
\label{app:sft_single}

\begin{figure}[h]
\centering
\includegraphics[width=\linewidth]{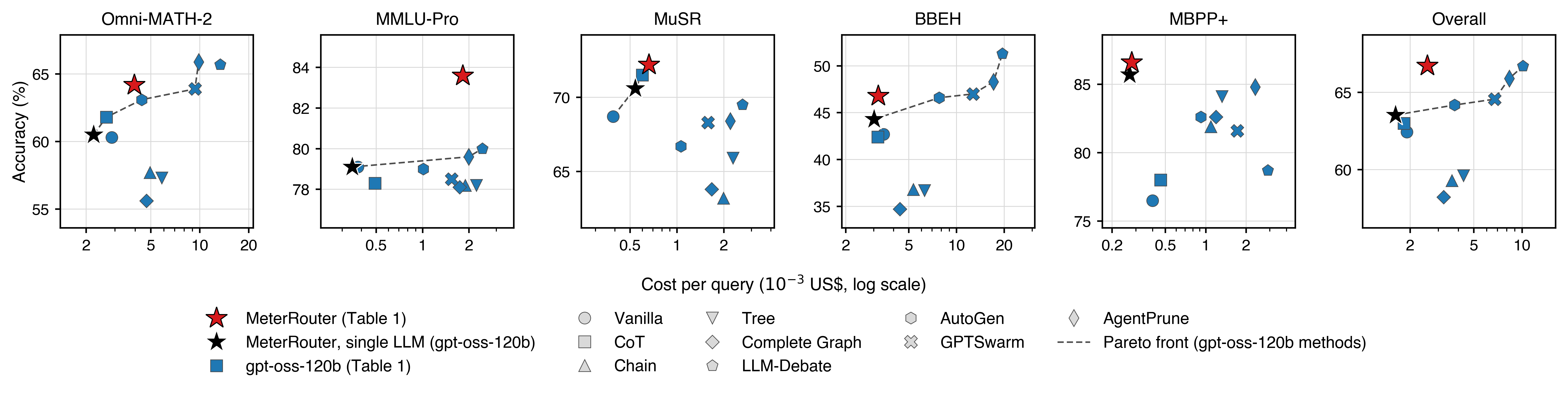} %
\caption{Accuracy versus real average cost per query for MeterRouter with a single LLM, the full MeterRouter, and the Table 1 baselines with gpt-oss-120b.}
\label{fig:sft_single}
\end{figure}

To separate graph construction from model routing, we train \texttt{MeterRouter} with one shared backbone, gpt-oss-120b, and leave the rest of the policy unchanged. The policy still builds the graph but can no longer choose models. Fig.~\ref{fig:sft_single} compares this variant with the full \texttt{MeterRouter} and with every baseline in Table~\ref{tab:baseline-results} that uses the same backbone, all under the budget $B=5\times10^{-2}$.

With one backbone, \texttt{MeterRouter} is 3.19 points below the full model overall and less accurate on every benchmark, but it keeps most of the gain. Learning the topology alone is enough to beat the same-backbone Chain, Tree, and Complete Graph baselines in Table~\ref{tab:baseline-results}: the single-LLM variant is more accurate and cheaper on every benchmark, and over all test questions it scores 3.92 to 5.31 points higher at half their cost or less. It also lies on the Pareto front of the gpt-oss-120b methods in all six panels. Four same-backbone multi-agent baselines are more accurate overall (AutoGen 64.18\%, GPTSwarm 64.56\%, AgentPrune 65.90\%, LLM-Debate 66.70\%), but each costs 2.3 to 6.3 times more per query.

\subsection{Fixed-Topology Ablation}
\label{app:fixed_topo}

\begin{figure}[h]
\centering
\includegraphics[width=\linewidth]{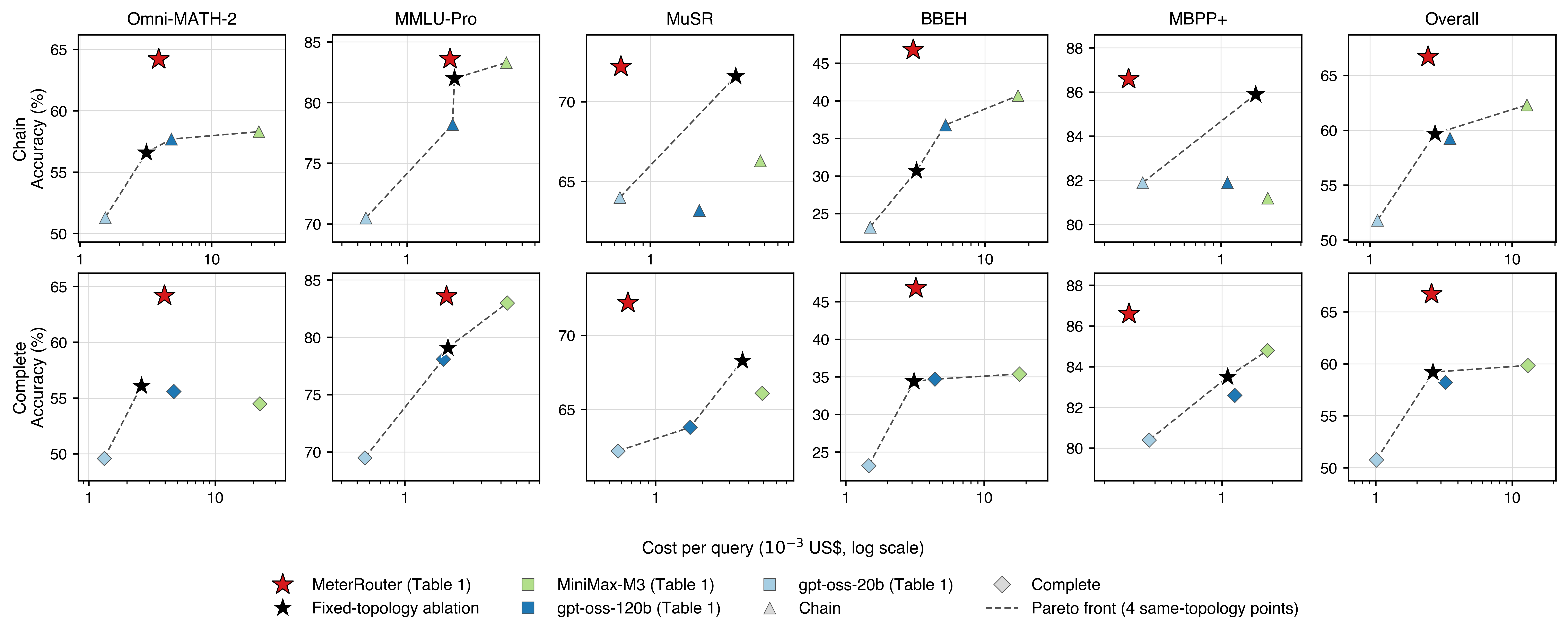} %
\caption{Accuracy versus cost per query for \texttt{MeterRouter} with a fixed topology (stars) and the baselines of the same topology, Chain (\textbf{top}) and Complete Graph (\textbf{bottom}).}
\label{fig:fixed_topo}
\end{figure}

To isolate the contribution of the topology, we fix \texttt{MeterRouter}'s topology to a chain or a complete graph and leave the rest of the policy unchanged. Fig.~\ref{fig:fixed_topo} compares each variant at $B=5\times10^{-2}$ with the baselines in Table~\ref{tab:baseline-results} that use the same topology. Budget-aware LLM allocation alone already improves a fixed-topology MAS: the fixed-topology variant lies on the Pareto frontier of the same-topology methods in all 12 panels. RL therefore allocates the heterogeneous LLMs efficiently. The full \texttt{MeterRouter} is still more accurate in every panel. Over all test questions, it scores 7.03 points higher than the chain variant and 7.51 points higher than the complete-graph variant, at a lower average cost.

\subsection{Budget Compliance and Termination}
\label{app:stop}

Each query ends by a learned stop, the agent cap, or a budget stop, and is graded normally. A query counts as incorrect if its final answer is truncated at its token cap or if no agent is affordable at all. An overrun is graded normally and counted as a violation. Table~\ref{tab:stop} reports how often each case occurs. 
\begin{itemize}
\item \textbf{Learned stop}: the policy's stop decision ends the graph. 
\item \textbf{Budget stop}: no backbone can afford another agent (Eq.~\ref{eq:mask}), so the meter ends the graph.
\item \textbf{Agent-cap stop}: the graph reaches $N_{\max}$ agents.
\item \textbf{Truncation}: the final answer is cut at its token cap (Eq.~\ref{eq:cap}).
\item \textbf{Overrun}: the realized cost exceeds $B$.
\end{itemize}

\begin{table}[h]
\centering
\caption{How queries end, and how often truncation and overruns occur, at each budget $B$. Learned, agent-cap and budget stops are the three ways a query ends and sum to 100\%.}
\label{tab:stop}
\resizebox{\linewidth}{!}{%
\begin{tabular}{lccccc}
\toprule
Budget & Learned stop & Agent-cap stop & Budget stop & Truncation & Overrun \\
\midrule
$B\in\{1,4\}\times10^{-3}$ & 86.77 $\pm$ 0.57 & 0.53 $\pm$ 0.20 & 12.70 $\pm$ 0.43 & 16.93 $\pm$ 0.34 & 0.02 $\pm$ 0.02 \\
$B=5\times10^{-3}$         & 91.06 $\pm$ 0.24 & 3.20 $\pm$ 0.12 & 5.74 $\pm$ 0.12  & 7.71 $\pm$ 0.24  & 0.01 $\pm$ 0.02 \\
$B=5\times10^{-2}$         & 88.25 $\pm$ 0.48 & 11.72 $\pm$ 0.46 & 0.03 $\pm$ 0.03 & 0.08 $\pm$ 0.12  & 0.00 $\pm$ 0.00 \\
\bottomrule
\end{tabular}}
\end{table}
\subsection{Adaptation to LLM Prices}
\label{app:price}

We test whether \texttt{MeterRouter} can adjust its LLM selection to changing prices through price-paired training, without changing the architecture or the budget meter. In each training round, one policy runs the same questions and budgets under three price multipliers ($0.1\times$, $1\times$, and $10\times$). The GRPO-style update pools the trajectories from all three settings and computes advantages within each question--budget--price group.

We conduct this experiment on MMLU-Pro by varying the price of
\texttt{gpt-oss-120b}. As shown in Table~\ref{tab:price}, after training, the router assigns gpt-oss-120b to the first agent in 92\% of queries at $0.1\times$, 23\% at $1\times$, and 0\% at $10\times$ (Table R4). As the model becomes more expensive and the router uses it less. Raising the price from $1\times$ to $10\times$ lowers spending, because the router moves away from the more expensive model.

\begin{table}[h]
\centering
\caption{Accuracy (\%) / cost per query ($\times10^{-3}$ USD) on MMLU-Pro at $B=5\times10^{-2}$ USD, after price-paired training, with the price of gpt-oss-120b scaled by each multiplier. First-choice rate is the share of queries whose first agent is assigned gpt-oss-120b.}
\label{tab:price}
\begin{tabular}{lcc}
\toprule
Price multiplier & \textbf{MMLU-Pro} & First-choice rate (\%) \\
\midrule
$0.1\times$ & 75.0 / 0.09 & 92.1 \\
$1\times$   & 68.2 / 0.21 & 23.2 \\
$10\times$  & 67.1 / 0.15 & 0.0  \\
\bottomrule
\end{tabular}
\end{table}

\subsection{Extending the LLM Pool}
\label{app:pool}

\begin{figure}[h]
\centering
\includegraphics[width=0.75\linewidth]{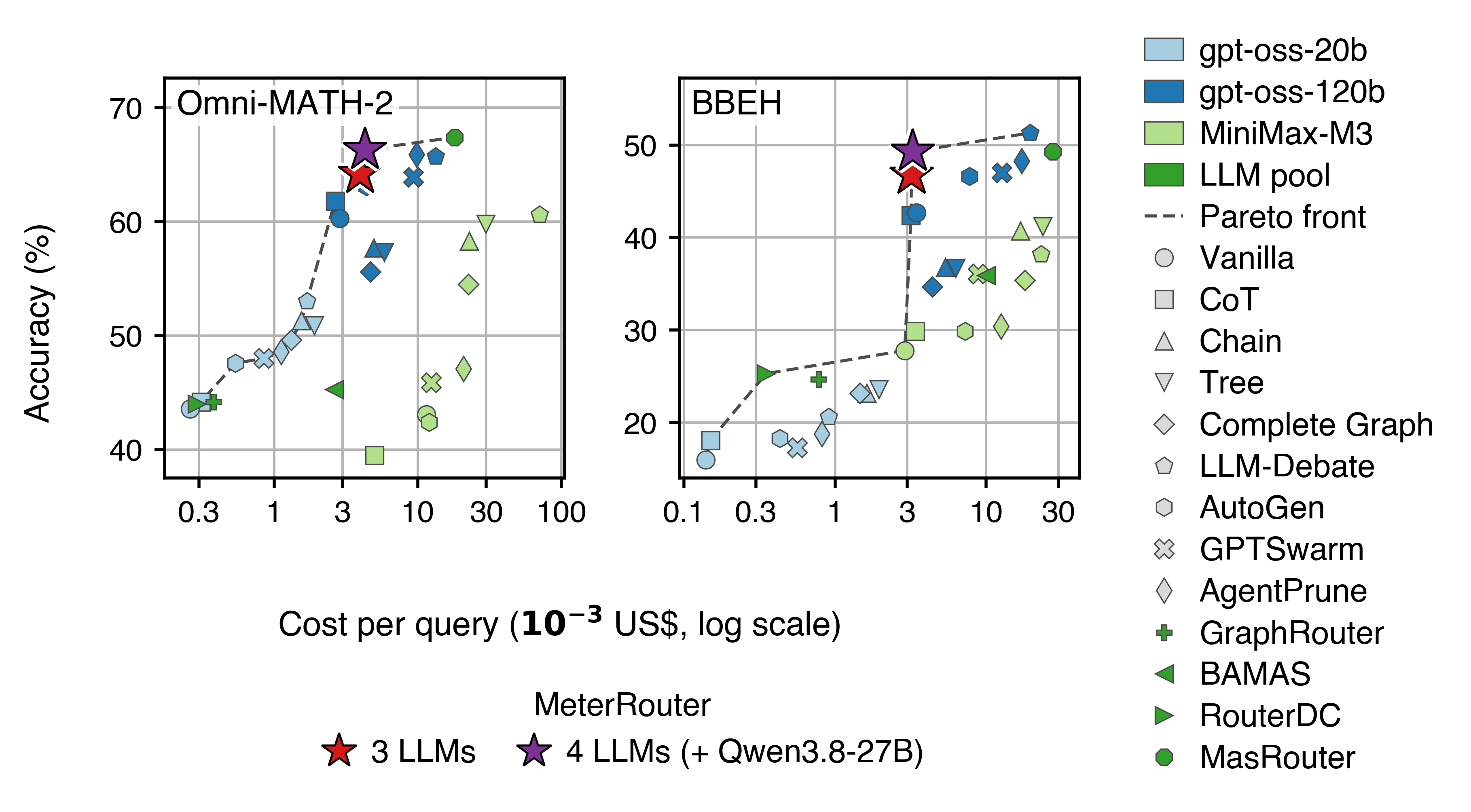} %
\caption{Accuracy versus cost per query on Omni-MATH-2 and BBEH for \texttt{MeterRouter} with the 3-LLM and 4-LLM pools (stars) and all baselines.}
\label{fig:pool}
\end{figure}

We add a fourth backbone, Qwen3.8-27B, as a new tier priced between gpt-oss-120b and MiniMax-M3 (\$0.20 and \$1.00 per million input and output tokens), and rerun the full pipeline (collection, SFT, and RL) on the 4-LLM pool without changing the architecture. We test on Omni-MATH-2 and BBEH, where \texttt{MeterRouter} has its lowest accuracy and which are the most constrained by the budget (Table~\ref{tab:ablation}). Accuracy rises by 2.1 points on Omni-MATH-2 and 2.5 points on BBEH, at 8\% and 2\% higher cost (Table~\ref{tab:pool}), and the router stays on the Pareto frontier of all methods (Fig.~\ref{fig:pool}).

\begin{table}[h]
\centering
\caption{Accuracy (\%) / cost per query ($\times10^{-3}$ \$) with the 3-LLM and 4-LLM pools at $B=5\times10^{-2}$. The 3-LLM row is from Table~\ref{tab:baseline-results}.}
\label{tab:pool}
\begin{tabular}{lcc}
\toprule
Pool & Omni-MATH-2 & BBEH \\
\midrule
3 LLMs & 64.2 / 3.95 & 46.8 / 3.20 \\
4 LLMs & 66.3 / 4.28 & 49.3 / 3.26 \\
\bottomrule
\end{tabular}
\end{table}

\section{Supplementary Details of MeterRouter}

\begin{table}[H]
\centering
\begin{minipage}{0.80\linewidth}
\centering
\caption{Templates in the predefined pool $\mathcal{T}$ used to build $\mathcal{D}_{\mathrm{SFT}}$. Templates below the rule are added for $N_{\max}=5$.}
\label{tab:graph_templates}
\begin{tabular}{lcc>{\raggedright\arraybackslash}p{0.52\linewidth}}
\toprule
\textbf{Template} &
\textbf{Agents} &
\textbf{Edges} &
\textbf{Parents per node} \\
\midrule

\texttt{single\_1}  & 1 & 0    & [] \\
\texttt{chain\_2}   & 2 & 1    & [], [0] \\
\texttt{chain\_3}   & 3 & 2    & [], [0], [1] \\
\texttt{fan-in\_3}   & 3 & 2    & [], [], [0, 1] \\
\texttt{dense\_3}   & 3 & 3    & [], [0], [0, 1] \\
\texttt{chain\_4}   & 4 & 3    & [], [0], [1], [2] \\
\texttt{fan-in\_4}   & 4 & 3    & [], [], [], [0, 1, 2] \\
\texttt{diamond\_4} & 4 & 4    & [], [0], [0], [1, 2] \\
\texttt{dense\_4}   & 4 & 6    & [], [0], [0, 1], [0, 1, 2] \\
\texttt{pruned\_4}  & 4 & 3--5 & Random per query \\
\texttt{chain\_5}   & 5 & 4    & [], [0], [1], [2], [3] \\
\texttt{fan-in\_5}   & 5 & 4    & [], [], [], [], [0, 1, 2, 3] \\
\texttt{dense\_5}   & 5 & 10   & [], [0], [0, 1], [0, 1, 2], [0, 1, 2, 3] \\
\texttt{pruned\_5}  & 5 & 5--9 & Random per query \\
\bottomrule
\end{tabular}

\end{minipage}
\end{table}

\section{Module Profile}
\label{app:role}
Each dataset in our experiments has its own role profile, written for that dataset's domain in the style of ARG-Designer~\cite{argdesigner}. All role profiles are summarized below.
\begin{roleprofile}{Omni-MATH-2}
role_profile = [
{
    'Name': 'Math Solver',
    'Description': "A math expert who solves the problem step by step, using other agents' answers only as hints.",
    'Prompt': 'You are a math expert. You will be given a math problem and hints from other agents. Give your own solving process step by step based on hints.'
    },
{
    'Name': 'Mathematical Analyst',
    'Description': 'A mathematical analyst who first analyzes the solution symbolically with letters, then substitutes values to calculate.',
    'Prompt': 'You are a mathematical analyst. You will be given a math problem, analysis and code from other agents. You need to first analyze the problem-solving process step by step, where the variables are represented by letters. Then you substitute the values into the analysis process to perform calculations and get the results.'
},
{
    'Name': 'Programming Expert',
    'Description': 'A programming expert who integrates step-by-step reasoning and Python code, which is executed to obtain the answer.',
    'Prompt': 'You are a programming expert. You will be given a math problem, analysis and code from other agents. Integrate step-by-step reasoning and Python code to solve math problems. Analyze the question and write code to solve the problem, using comments for the reasoning.',
},
{
    'Name': 'Inspector',
    'Description': 'An inspector who checks the logic, calculations, and code of other agents and gives a verified solution.',
    'Prompt': 'You are an Inspector. You will be given a math problem, analysis and code from other agents. Check whether the logic/calculation of the problem solving and analysis process is correct (if present). Check whether the code corresponds to the solution analysis (if present). Give your own solving process step by step based on hints.'
}
]
\end{roleprofile}

\begin{roleprofile}{MBPP+}
role_profile = [
{
    'Name': 'Project Manager',
    'Description': 'A project manager who oversees the overall code structure and suggests a concise, maintainable design.',
    'Prompt': 'You are a project manager. You will be given a function description and an example test by the user. You are responsible for overseeing the overall structure of the code, ensuring that the code is structured to complete the task. Implement code concisely and correctly without pursuing over-engineering. You need to suggest optimal design patterns to ensure that the code follows best practices for maintainability and flexibility. You can specify the overall design of the code, including the classes that need to be defined (maybe none) and the functions used (maybe only one function).'
},
{
    'Name': 'Algorithm Designer',
    'Description': 'An algorithm designer who specifies the algorithm and, for complex logic, its pseudocode.',
    'Prompt': 'You are an algorithm designer. You will be given a function description and an example test by the user. You need to specify the specific design of the algorithm, including the classes that may be defined and the functions used. When the implementation logic is complex, you can give the pseudocode logic of the main algorithm.'
},
{
    'Name': 'Programming Expert',
    'Description': "A programming expert who writes the full implementation, considering other agents' code and test results.",
    'Prompt': 'You are a programming expert. You will be given a function description and an example test by the user. You may be able to get the output results of other agents. They may have passed the example test, but they may not be completely correct. Write your full implementation (restate the function signature). Do not include anything other than a Python code block in your response. Do not change function names and input variable types in tasks.'
},
{
    'Name': 'Test Analyst',
    'Description': 'A test analyst who identifies problems in current code, special cases, and boundary conditions.',
    'Prompt': 'You are a test analyst. You will be given a function description and an example test by the user. You need to provide problems in the current code or solution based on the test data and possible test feedback in the question. You need to provide additional special use cases, boundary conditions, etc. that should be paid attention to when writing code. You can point out any potential errors in the code.'
},
{
    'Name': 'Bug Fixer',
    'Description': 'A bug fixer who provides modified and improved code based on the design, implementation, and test problems.',
    'Prompt': 'You are a bug fixer. You will be given a function description and an example test by the user. You need to provide modified and improved python code based on the current overall code design, algorithm framework, code implementation or test problems. Write your full implementation (restate the function signature). Do not include anything other than a Python code block in your response. Do not change function names and input variable types in tasks.'
},
{
    'Name': 'Normal Programmer',
    'Description': "A programmer who writes a full implementation, referring to other agents' outputs.",
    'Prompt': "You are a programmer. You will be given a function description and an example test by the user. You can refer to the agents' outputs. Write your full implementation (restate the function signature). Do not include anything other than a Python code block in your response."
}
]
\end{roleprofile}

\begin{roleprofile}{MMLU-Pro}
role_profile = [
{
    'Name': 'Knowledgeable Expert',
    'Description': 'A knowledgeable expert in question answering who identifies the key entities and the facts needed to answer.',
    'Prompt': 'You are a knowledgeable expert in question answering. Identify the key entities and concepts that the question depends on, and state the relevant facts you know about them before choosing an answer.'
},
{
    'Name': 'Critic',
    'Description': "An excellent critic who points out potential issues in other agents' analysis point by point.",
    'Prompt': "You are an excellent critic. Please point out potential issues in other agent's analysis point by point."
},
{
    'Name': 'Mathematician',
    'Description': 'A mathematician good at math games, arithmetic calculation, and long-term planning.',
    'Prompt': 'You are a mathematician who is good at math games, arithmetic calculation, and long-term planning.'
},
{
    'Name': 'Psychologist',
    'Description': 'A psychologist good at psychology, sociology, and philosophy.',
    'Prompt': 'You are a psychologist. You are good at psychology, sociology, and philosophy, and you reason scientifically about human behavior.'
},
{
    'Name': 'Historian',
    'Description': 'A historian who researches and analyzes cultural, economic, political, and social events in the past.',
    'Prompt': 'You research and analyze cultural, economic, political, and social events in the past, collect data from primary sources and use it to develop theories about what happened during various periods of history.'
},
{
    'Name': 'Doctor',
    'Description': 'A doctor with knowledge of medicine, physiology, and health.',
    'Prompt': "You are a doctor. You are knowledgeable about medicine, physiology, diseases, and their treatment, and you consider a patient's age, lifestyle, and medical history when reasoning about health questions."
},
{
    'Name': 'Lawyer',
    'Description': 'A lawyer good at law, politics, and history.',
    'Prompt': 'You are good at law, politics, and history.'
},
{
    'Name': 'Economist',
    'Description': 'An economist good at economics, finance, and business.',
    'Prompt': 'You are good at economics, finance, and business. You have experience on understanding charts while interpreting the macroeconomic environment prevailing across world economies.'
},
{
    'Name': 'Programmer',
    'Description': 'A programmer good at computer science, engineering, and physics.',
    'Prompt': 'You are good at computer science, engineering, and physics. You have experience in designing and developing computer software and hardware.'
}
]
\end{roleprofile}

\begin{roleprofile}{MuSR}
role_profile = [
{
    'Name': 'Evidence Tracker',
    'Description': "An evidence tracker who lists the story's facts about means, motive, opportunity, and alibis, separating stated facts from assumptions.",
    'Prompt': 'You are an evidence tracker. You will be given a long story, a question with options, and outputs from other agents. Extract the facts in the story that bear on the question, such as means, motive, opportunity, alibis, and contradictions. Separate what the story states from what is assumed, then weigh the evidence for each option step by step.'
},
{
    'Name': 'Timeline and Location Analyst',
    'Description': 'A timeline and location analyst who tracks where objects and people are over time and who saw each move.',
    'Prompt': "You are a timeline and location analyst. You will be given a long story, a question with options, and outputs from other agents. Track step by step where each object and person is over time, who moved what, and who witnessed or missed each move, so that each character's beliefs about locations are clear. Then answer the question."
},
{
    'Name': 'Constraint Planner',
    'Description': "A constraint planner who turns people's skills and relationships into constraints and evaluates each assignment.",
    'Prompt': "You are a constraint planner. You will be given a long story, a question with options, and outputs from other agents. List each person's relevant skills, weaknesses, and working relationships as constraints, then evaluate every option against those constraints step by step and choose the assignment that best satisfies them."
},
{
    'Name': 'Critic',
    'Description': "An excellent critic who points out unsupported inferences and overlooked story details in other agents' analysis.",
    'Prompt': "You are an excellent critic. You will be given a long story, a question with options, and outputs from other agents. Please point out unsupported inferences, overlooked story details, and contradictions in other agents' analysis point by point."
},
{
    'Name': 'Inspector',
    'Description': "An inspector who checks each option against the story's evidence and other agents' reasoning, then gives a verified answer.",
    'Prompt': "You are an Inspector. You will be given a long story, a question with options, and outputs from other agents. Check whether other agents' reasoning is supported by the story, check each option against the decisive evidence, and give your own verified reasoning step by step."
}
]
\end{roleprofile}

\begin{roleprofile}{BBEH}
role_profile = [
{
    'Name': 'Math Solver',
    'Description': "A solver who works through the task step by step, using other agents' answers only as hints.",
    'Prompt': 'You are a math expert. You will be given a hard reasoning task and hints from other agents. Give your own solving process step by step based on hints.'
},
{
    'Name': 'Mathematical Analyst',
    'Description': "A mathematical analyst who analyzes the task's rules symbolically, then substitutes values to calculate step by step.",
    'Prompt': "You are a mathematical analyst. You will be given a hard reasoning task and outputs from other agents. You need to first analyze the task's rules and the problem-solving process step by step, where unknown quantities are represented by letters. Then you substitute the values into the analysis process to perform calculations and get the results."
},
{
    'Name': 'Programming Expert',
    'Description': "A programming expert who encodes the task's rules in Python code, which is executed to obtain the answer.",
    'Prompt': "You are a programming expert. You will be given a hard reasoning task and outputs from other agents. Encode the task's rules, operations, and data faithfully in Python and compute the result, using comments for the reasoning."
},
{
    'Name': 'Inspector',
    'Description': "An inspector who checks whether other agents followed the task's exact rules and answer format, then gives a verified answer.",
    'Prompt': "You are an Inspector. You will be given a hard reasoning task and outputs from other agents. Check whether the logic, calculations, and code of other agents follow the task's exact rules (if present), and whether their answers use the representation the task asks for. Give your own solving process step by step based on hints."
},
{
    'Name': 'Logician',
    'Description': 'A logician who formalizes clues and constraints, deduces consequences, and eliminates inconsistent possibilities.',
    'Prompt': 'You are a logician. You will be given a hard reasoning task and outputs from other agents. Formalize the clues, statements, and constraints, deduce their consequences step by step, and eliminate possibilities that contradict them until the answer is determined.'
},
{
    'Name': 'Linguist',
    'Description': 'A linguist who analyzes word order, meaning, ambiguity, humor, and unfamiliar language patterns.',
    'Prompt': 'You are a linguist. You will be given a hard reasoning task and outputs from other agents. Analyze the language in the task, including grammar, word order, word meanings, ambiguity, sarcasm or humor, and patterns in unfamiliar languages, and reason step by step to the answer.'
},
{
    'Name': 'Critic',
    'Description': "An excellent critic who points out rule violations and reasoning errors in other agents' analysis.",
    'Prompt': "You are an excellent critic. You will be given a hard reasoning task and outputs from other agents. Please point out rule violations, reasoning errors, and answer-format mistakes in other agents' analysis point by point."
}
]
\end{roleprofile}

\section{Configuration of Baseline Methods}
\label{app:baseline}

\begin{table*}[ht]
\centering
\newcommand{\stk}[1]{\begin{tabular}[t]{@{}l@{}}#1\end{tabular}}
\caption{Experimental settings of the baselines. \textbf{Calls/Query} is the average number of LLM calls per evaluated question.}
\label{tab:baseline_set}
\renewcommand{\arraystretch}{1.1}
\resizebox{\textwidth}{!}{%
\begin{tabular}{lllcl}
\toprule
\rowcolor{gray!35}
\textbf{Method} &
\textbf{Training} &
\textbf{Structure} &
\textbf{Calls/Query} &
\textbf{Hyper-params.} \\
\midrule

\rowcolor{gray!12}
Vanilla
& None
& \stk{Single LLM,\\ zero-shot}
& 1
& -- \\

CoT
& None
& \stk{Single LLM,\\ 3-shot CoT}
& 1
& -- \\

\midrule

\rowcolor{gray!12}
Chain
& None
& 5 agents, chain
& 5
& No pruning \\

Tree
& None
& \stk{5 agents, binary tree\\ + sink aggregation}
& 6
& No pruning \\

\rowcolor{gray!12}
Complete Graph
& None
& \stk{5 agents,\\ complete DAG}
& 5
& No pruning \\

LLM-Debate
& None
& \stk{3 agents, 2 rounds,\\ majority vote}
& 6
& No pruning \\
\midrule
\rowcolor{gray!12}
AutoGen
& None
& \stk{Round-robin team:\\ solver, critic, solver}
& 3
& Max 3 turns \\

GPTSwarm
& None
& \stk{3 IO agents +\\ final decision node}
& 4
& \stk{Fixed graph (no edge/\\ node optimization)} \\

\rowcolor{gray!12}
AgentPrune
& \stk{Edge masks learned on\\ the training split}
& \stk{5 agents, fully connected,\\ + decision node; pruned}
& 6
& \stk{1 round (spatial edges only);\\ 10 iters, batch 4, lr 0.1;\\ prune 25\% every 5 iters} \\

\midrule

GraphRouter
& \stk{GNN router on MiniLM\\ embeddings, 1000 epochs;\\ labels from each LLM}
& \stk{One LLM\\ per query}
& 1
& \stk{Performance First;\\ batch 32, lr $3{\times}10^{-4}$} \\

\rowcolor{gray!12}
RouterDC
& \stk{mDeBERTa-v3-base router,\\ 1000 steps; labels from\\ 10 samples/LLM ($T{=}0.2$)}
& \stk{One LLM\\ per query}
& 1
& \stk{Batch 64, lr $5{\times}10^{-5}$;\\ $K^{+}{=}1$, $K^{-}{=}2^{\S 2}$;\\ 5 clusters} \\

BAMAS
& \stk{Per-benchmark policy,\\ $0.05$ budget, 10 epochs}
& \stk{4 candidate topologies;\\ ILP team of 5 (10 on Omni$^{\S 1}$\\ due to different topology)}
& 3.9
& \stk{Budget \$0.05/query;\\ linear $\le5$ steps} \\

\rowcolor{gray!12}
MasRouter
& \stk{Per-benchmark controller,\\ 5 epochs (10 on MBPP+)}
& \stk{Up to 6 agents,\\ 6 collab.\ modes}
& 9.0
& \stk{Batch 16, lr 0.01; cost\\ penalty 100 (Omni$^{\S 1}$),\\ 400 (MBPP+), 200 (rest)} \\

\bottomrule
\end{tabular}%
}
\par\smallskip
\begin{minipage}{\textwidth}
\footnotesize
$^{\S 1}$Omni denotes Omni-Math-2. $^{\S 2}$The paper uses $K^{+}=K^{-}=3$, which is degenerate with three candidate LLMs.
\end{minipage}
\end{table*}

\end{document}